\documentclass[11pt]{article}

\usepackage[final]{acl}

\usepackage{times}
\usepackage{latexsym}
\usepackage{amsmath}
\usepackage{amssymb}
\usepackage{xcolor}
\usepackage{tikz}
\usepackage[most]{tcolorbox}
\usepackage{caption}
\usepackage[T1]{fontenc}

\usepackage[utf8]{inputenc}
\usepackage{booktabs}   
\usepackage{multirow}   
\usepackage{amsmath}
\usepackage{algorithm}
\usepackage{algpseudocode}
\usepackage{microtype}
\usepackage{booktabs}
\usepackage{inconsolata}

\usepackage{graphicx}
\usepackage{float} 
\newcommand{\geminilogo}{\raisebox{-0.15ex}{\includegraphics[height=0.9em]{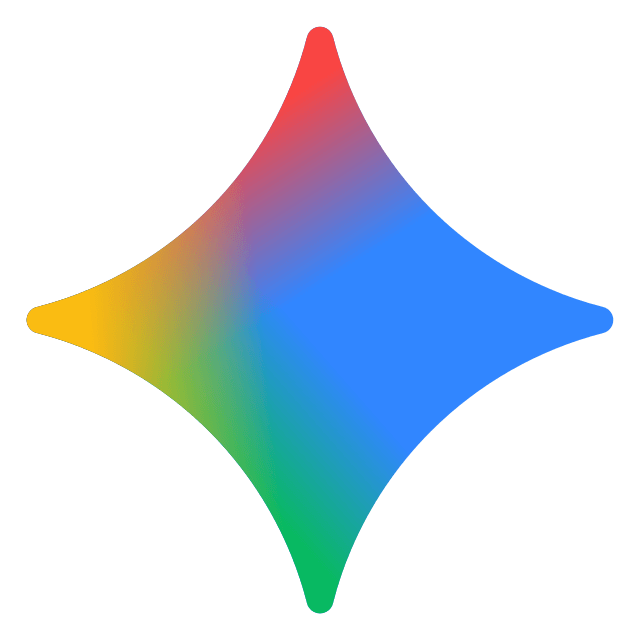}}\hspace{2pt}}
\newcommand{\audioflamingologo}{\raisebox{-0.15ex}{\includegraphics[height=0.9em]{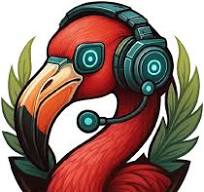}}\hspace{2pt}}
\newcommand{\gemmalogo}{\raisebox{-0.15ex}{\includegraphics[height=0.9em]{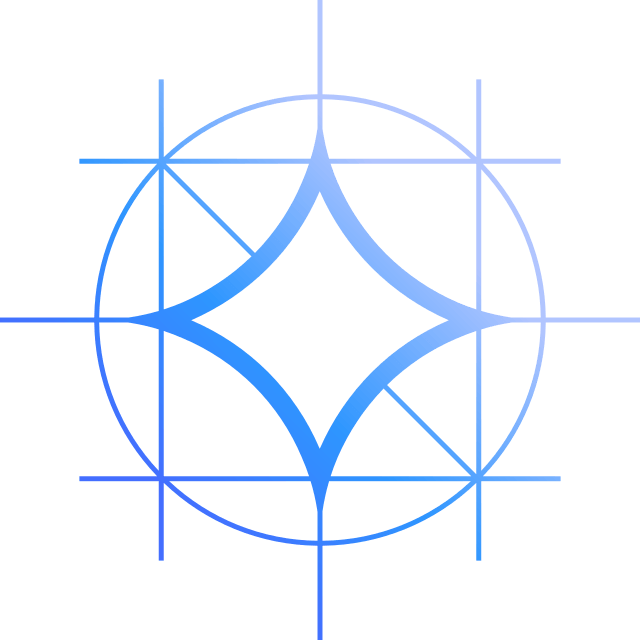}}\hspace{2pt}}
\newcommand{\openailogo}{\raisebox{-0.15ex}{\includegraphics[height=0.9em]{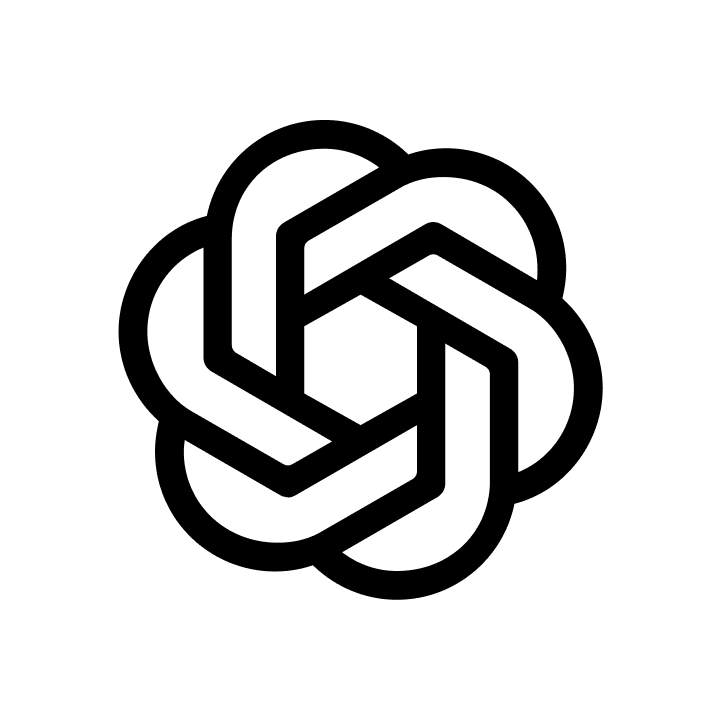}}\hspace{2pt}}
\newcommand{\kimilogo}{\raisebox{-0.15ex}{\includegraphics[height=0.9em]{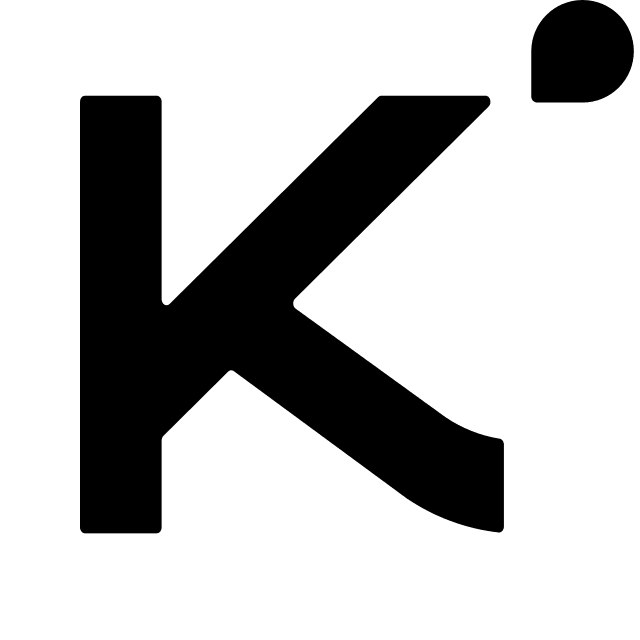}}\hspace{2pt}}
\newcommand{\microsoftlogo}{\raisebox{-0.15ex}{\includegraphics[height=0.9em]{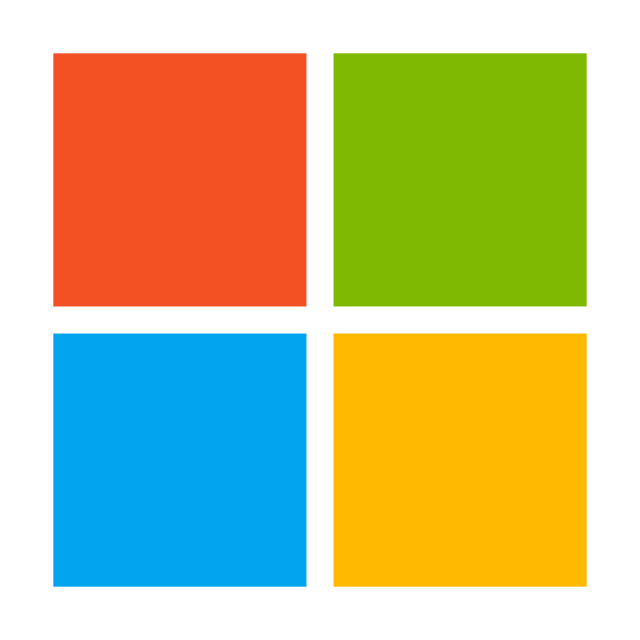}}\hspace{2pt}}
\newcommand{\qwenlogo}{\raisebox{-0.15ex}{\includegraphics[height=0.9em]{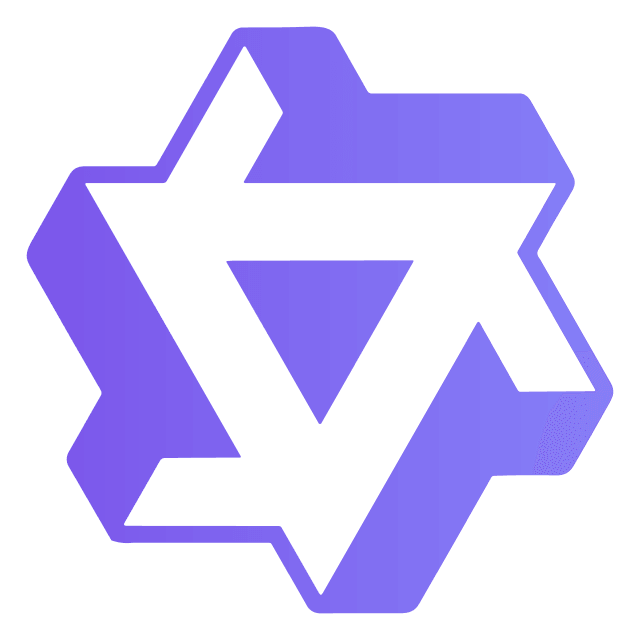}}\hspace{2pt}}
\newcommand{\openbmblogo}{\raisebox{-0.15ex}{\includegraphics[height=0.9em]{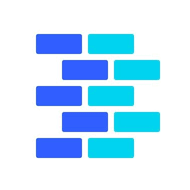}}\hspace{2pt}}
\newcommand{\mistrallogo}{\raisebox{-0.15ex}{\includegraphics[height=0.9em]{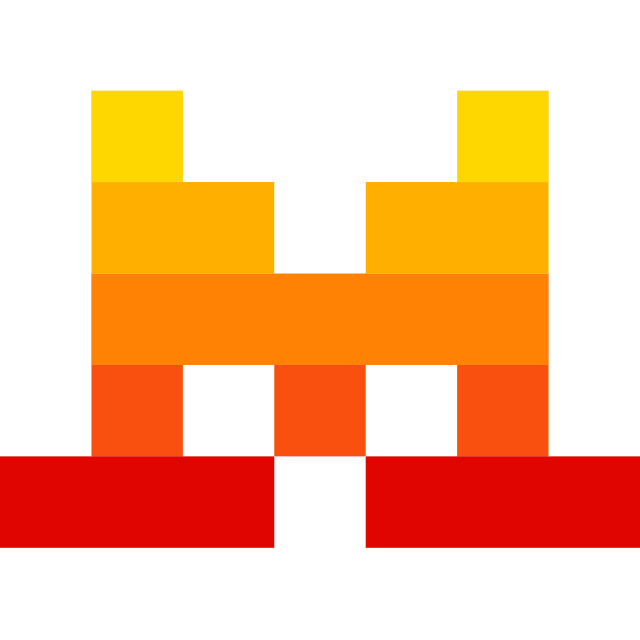}}\hspace{2pt}}
\newcommand{\glmlogo}{\raisebox{-0.15ex}{\includegraphics[height=0.9em]{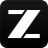}}\hspace{2pt}}
\newcommand{\stepfunlogo}{\raisebox{-0.15ex}{\includegraphics[height=0.9em]{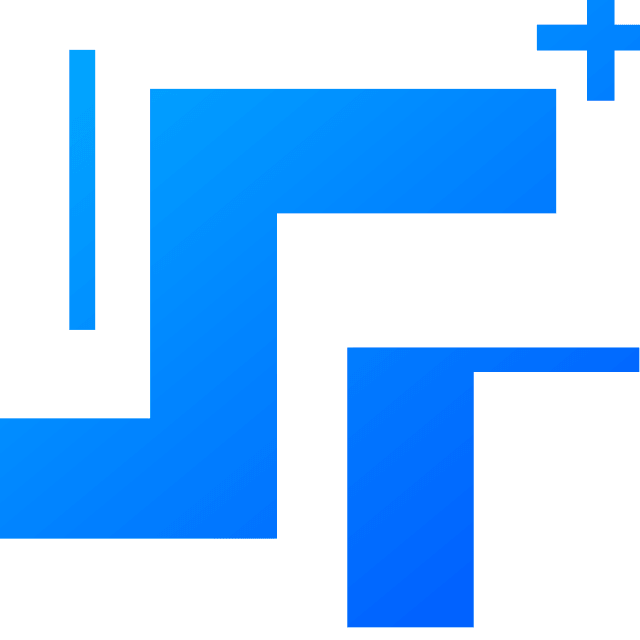}}\hspace{2pt}}
\definecolor{pastelblue}{RGB}{120,180,255}

\definecolor{boxbg}{HTML}{F8F9FA}       
\definecolor{boxframe}{HTML}{E9ECEF}    
\definecolor{textmain}{HTML}{212529}    
\definecolor{textmuted}{HTML}{6C757D}   
\definecolor{accentDiscr}{HTML}{3B82F6} 
\definecolor{accentAttr}{HTML}{10B981}  
\definecolor{accentReas}{HTML}{8B5CF6}  
\definecolor{audiobg}{HTML}{E2E8F0}
\definecolor{audiotext}{HTML}{334155}
\definecolor{slotbg}{HTML}{F1F5F9}
\definecolor{slottext}{HTML}{0F172A}

\newtcolorbox{promptbox}[2]{%
  enhanced, breakable=false,
  colback=boxbg, colframe=boxframe,
  boxrule=0.5pt, arc=3pt, outer arc=3pt,
  leftrule=3.5pt,
  borderline west={3.5pt}{0pt}{#1},
  left=10pt, right=10pt, top=8pt, bottom=8pt,
  width=\textwidth,
  before skip=4pt, after skip=4pt,
  fontupper=\sffamily\small\color{textmain},
  overlay={
    \node[anchor=north east, yshift=-4pt, xshift=-4pt] at (frame.north east)
    {\tcbox[enhanced, colback=#1!10, colframe=#1!20, boxrule=0.5pt, arc=2pt,
            left=4pt, right=4pt, top=1.5pt, bottom=1.5pt, nobeforeafter]
      {\sffamily\bfseries\scriptsize\textcolor{#1}{#2}}};
  }
}

\newcommand{\audiobadge}[1]{%
  \tcbox[%
    enhanced, on line,
    colback=audiobg, colframe=audiobg,
    boxrule=0pt, arc=2pt,
    left=5pt, right=5pt, top=2.5pt, bottom=2.5pt,
    nobeforeafter, box align=base,
    fontupper=\sffamily\scriptsize\bfseries\color{audiotext},
  ]{\raisebox{-0.3pt}{$\blacktriangleright$}\ \,#1}%
  \par\vspace{5pt}%
}

\newcommand{\slot}[1]{%
  \tcbox[%
    enhanced, on line,
    colback=slotbg, colframe=slotbg!80!black,
    boxrule=0.3pt, arc=2pt,
    left=2pt, right=2pt, top=1pt, bottom=1pt,
    nobeforeafter, box align=base,
  ]{\texttt{\textbf{\textcolor{slottext}{\ensuremath{\langle #1 \rangle}}}}}%
}
\newcommand{\slotmath}[1]{%
  \tcbox[%
    enhanced, on line,
    colback=slotbg, colframe=slotbg!80!black,
    boxrule=0.3pt, arc=2pt,
    left=2pt, right=2pt, top=1pt, bottom=1pt,
    nobeforeafter, box align=base,
  ]{\texttt{\textbf{\textcolor{slottext}{$\langle #1 \rangle$}}}}%
}
\newcommand{\taskcell}[6]{%
  \begin{minipage}[t]{\textwidth}
    \begin{promptbox}{#1}{#2}
      \noindent\textcolor{textmuted}{\sffamily\scriptsize\MakeUppercase{#3}}\par\vspace{4pt}
      #4
      \vspace{4pt}
      \begin{tcolorbox}[enhanced, breakable=false, colback=white, colframe=boxframe,
                        boxrule=0.5pt, arc=2pt, left=8pt, right=8pt, top=6pt, bottom=6pt]
        #5
        \vspace{4pt}\hrule height 0.3pt\vspace{4pt}
        \textcolor{textmuted}{\scriptsize #6}
      \end{tcolorbox}
    \end{promptbox}
  \end{minipage}%
}

\title{HEAR Who Said What: Unlocking Speaker-Attributed Reasoning via Counterfactual Voice Grounding}

\author{
\textbf{Dongwook Lee\textsuperscript{1}} \quad 
\textbf{Sangkwon Park\textsuperscript{2}} \quad
\textbf{Eunwoo Song\textsuperscript{3}} \quad
\textbf{Che Hyun Lee\textsuperscript{2}} 
\\
\textbf{Youngho Cho\textsuperscript{1}} \
\textbf{Junho Kim\textsuperscript{1}} \
\textbf{June Young Yi\textsuperscript{4}} \
\textbf{Heeseung Kim\textsuperscript{5}}\footnotemark[2] \
\textbf{Sungroh Yoon\textsuperscript{1,2,6}}\footnotemark[2]
\\
\footnotetext[2]{Corresponding authors.}
\\
\textsuperscript{1}IPAI, Seoul National University (SNU) \quad \textsuperscript{2}Department of ECE, SNU \\
\textsuperscript{3}Department of EE, Yonsei University \quad \textsuperscript{4}Department of CSE, SNU \\
\textsuperscript{5}Department of AI, University of Seoul \quad \textsuperscript{6}AIIS, ASRI, INMC, and ISRC, SNU
}

\begin{document}
\maketitle

\renewcommand{\thefootnote}{\fnsymbol{footnote}} 
\footnotetext[2]{Corresponding authors.}

\begin{abstract}
Speech Language Models (SLMs) are increasingly deployed in multi-speaker environments, yet their ability to attribute speech to the correct speaker and reason over speaker identities remains unclear. Hence, we introduce \textbf{HEAR}, a conceptually hierarchical benchmark diagnosing the foundational capabilities of speaker-attributed reasoning, comprising 2.4K human-verified samples from 887 diverse multi-party audio clips. Evaluating 20 leading SLMs on HEAR reveals they struggle with these foundational tasks, often relying on semantic priors rather than actual vocal cues. To address this, we present \textbf{A2R}, a 30B model optimized on \textit{Counterfactual Audio with Speaker-level Hard negatives} (\textbf{CASH}), a dataset designed to guide the model to prioritize acoustic vocal cues over linguistic signals. A2R achieves strong performance on HEAR and exhibits zero-shot generalization to diverse multi-speaker downstream tasks, demonstrating that learned speaker attribution unlocks the model's latent capacity for speaker-aware reasoning. All resources are available on our \href{https://attributetoreason.github.io/AttributeToReason/}{\texttt{\textcolor{pastelblue}{project page}}}.
\end{abstract}

\section{Introduction}
Understanding speech in real-world environments often requires handling multi-speaker scenarios, where multiple voices are present and interact dynamically~\citep{lee2026usevaluatingimprovingvoice}. Identifying who said what, a task commonly referred to as speaker attribution~\cite{kanda2021comparativestudymodularjoint}, is a fundamental prerequisite for any higher-level understanding or reasoning in multi-party scenarios. This challenge becomes critical as speech language models (SLMs) are increasingly deployed in multi-party environments~\citep{nguyen2025seehearunderstandbenchmarking}. However, prior works~\cite{kumar2025mmauprochallengingcomprehensivebenchmark, sakshi2024mmaumassivemultitaskaudio} often evaluate speaker attribution only through its downstream tasks, without directly testing the foundational abilities required to support it. This makes it difficult to localize where models actually fail: whether they struggle to discriminate between distinct voices, attribute utterances to the correct speakers, or simply lack the reasoning competency for the downstream task.

To address this gap, we introduce a benchmark, \textbf{HEAR}, designed to evaluate the foundational abilities underlying speaker-attributed reasoning in SLMs. HEAR is grounded in Erber’s Auditory Hierarchy~\cite{alma991028407999703276}, which organizes human auditory understanding as a progression from perceptual discrimination to identification and higher-level comprehension. It maps Erber's view into three dimensions: \textbf{(i) Discrimination}, the ability to distinguish distinct voice identities; \textbf{(ii) Attribution}, the capacity to bind linguistic content to the exact voice identity; and \textbf{(iii) Reasoning}, the capacity to perform basic reasoning over speaker-attributed utterances. By mirroring this human cognitive progression, we provide a biologically inspired taxonomy to diagnose whether SLMs' failures stem from low-level acoustic perception or high-level semantic reasoning. The benchmark consists of 887 real-world audio clips and 2.4K human-verified multiple-choice question-answer pairs, covering diverse real-world acoustic settings such as daily conversations, meetings, and documentaries.

Our evaluations on the HEAR benchmark expose vulnerabilities in most leading SLMs. These models fail at basic attribution and exhibit an over-reliance on semantic priors, ignoring acoustic vocal cues when semantic context is available. To mitigate this semantic over-reliance, we introduce \textit{Counterfactual Audio with Speaker-level Hard negatives} \textbf{(CASH)}, a 60K-scale corpus spanning all task dimensions of HEAR, designed to decouple acoustic identity from semantic content. By using voice cloning to swap speakers for specific utterances while preserving the original transcript (Figure~\ref{fig:voice_swap}-(A)), CASH yields hard negatives where the correct answer flips solely based on vocal cues, thereby compelling models to ground their predictions in acoustic evidence rather than textual priors. 

Building on this dataset, we present \textbf{A2R} (Attribution-to-Reasoning), a speech language model optimized from Qwen3-Omni-30B-A3B-Instruct~\cite{xu2025qwen3omnitechnicalreport} via Group Relative Policy Optimization (GRPO)~\cite{shao2024deepseekmathpushinglimitsmathematical} to strengthen speaker attribution while preserving the base model's strong inherent reasoning ability. It is trained under a transcription-first objective: it first produces a speaker-tagged transcript and then reasons over it.

Consequently, A2R not only excels on HEAR but also demonstrates robust zero-shot transfer across various downstream multi-speaker benchmarks that require speaker attribution. These results suggest that speaker attribution is not merely an auxiliary sub-task, but a key mechanism for resolving the acoustic bottleneck in multi-party auditory comprehension. By explicitly forcing the model to ground utterances in vocal cues, we show that speaker-attributed reasoning can be elicited from the model in multi-speaker environments.

Our contributions can be summarized as follows: \textbf{(1)} We present \textbf{HEAR}, a benchmark comprising 2.4K human-annotated samples, designed to evaluate the foundational capabilities for speaker-attributed reasoning. \textbf{(2)} We reveal that current leading SLMs exhibit \emph{semantic hallucination} in speaker-attributed reasoning, exploiting language priors rather than the voice cues that actually determine who said what. \textbf{(3)} We propose \textit{Counterfactual Audio with Speaker-level Hard negatives} (\textbf{CASH}), a 60K-scale dataset designed to guide models to prioritize vocal cues over semantic content. \textbf{(4)} We introduce \textbf{A2R}, a 30B model trained with a transcription-first objective that rewards accurate transcriptions before reasoning, leading to strong improvements on HEAR and transfer gains on benchmarks requiring speaker attribution.

\begin{figure*}[h]
    \centering
    \includegraphics[width=\textwidth]{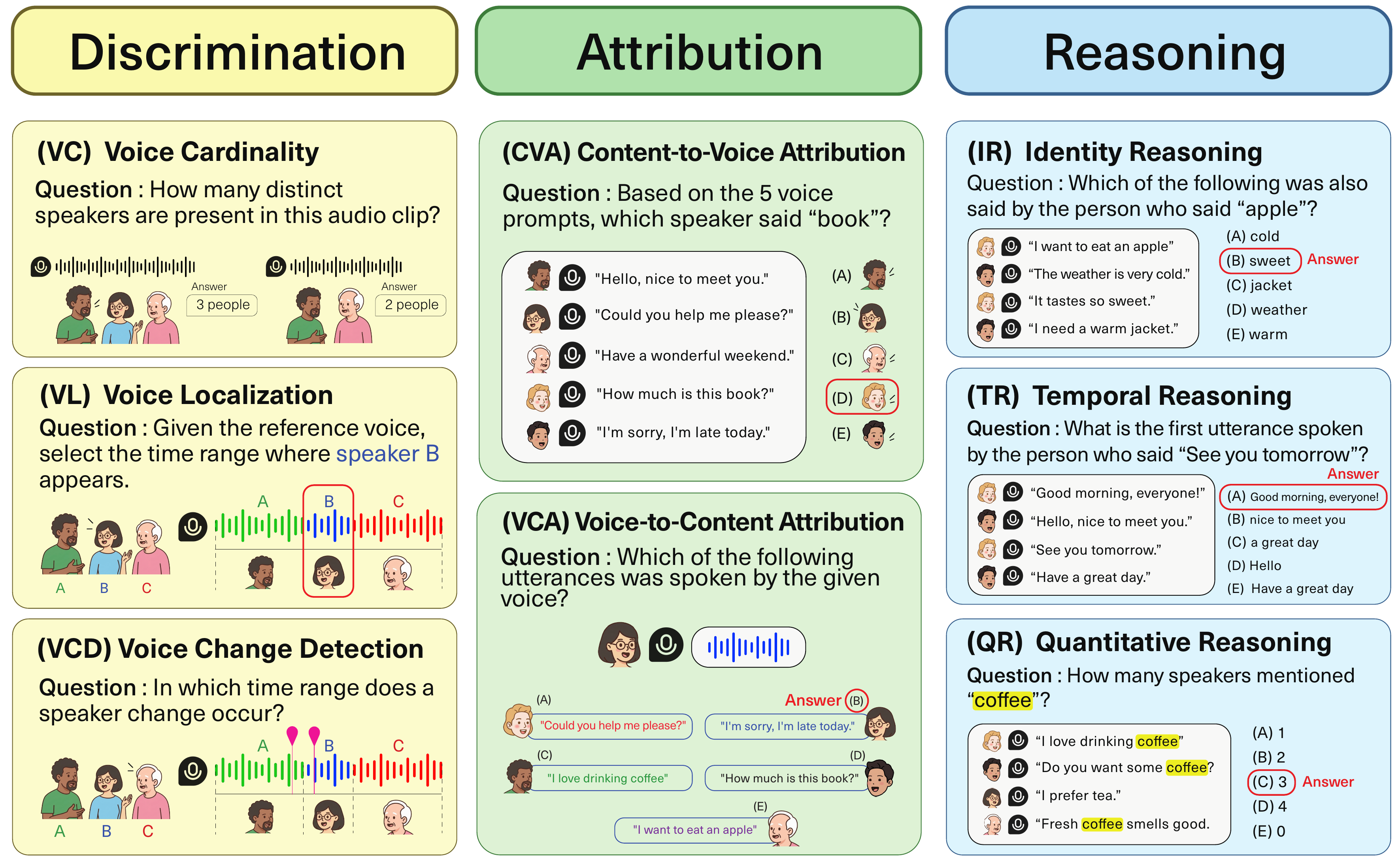}
    \caption{\textbf{Illustration of HEAR benchmark.} Inspired by Erber's auditory hierarchy~\cite{alma991028407999703276}, HEAR breaks speaker-attributed reasoning down into several foundational capabilities: discrimination, attribution, and reasoning.}
    \label{fig:illustration_hear}
\end{figure*}

\section{HEAR Benchmark}
\label{benchmark}
In this section, we introduce \textbf{HEAR}, a benchmark for Hierarchical Evaluation of Attribution and Reasoning, designed to evaluate the foundational capabilities underlying speaker-attributed reasoning in SLMs. The benchmark is formatted as Multiple-Choice Question Answering (MCQA) and organized into a hierarchy that mirrors human auditory processing~\cite{alma991028407999703276}, as illustrated in Figure~\ref{fig:illustration_hear}: (i) Discrimination, (ii) Attribution, and (iii) Reasoning. To analyze the impact of overlapping speech, we distinguish cases based on whether the overlapping utterance segment is needed to answer the question (Appendix~\ref{app:overlap_labeling_Criteria}).

\subsection{Taxonomy}
\subsubsection{Discrimination}
This axis evaluates the capacity to discriminate different voice identities and localize their temporal boundaries.

\paragraph{Voice Cardinality (VC)} This task evaluates the model's ability to determine the total number of unique speakers present within a given audio clip. Representing the most fundamental level of speaker-identity clustering, VC assesses whether the model can accurately quantify the underlying voice population of an acoustic scene. 

\paragraph{Voice Localization (VL)} Given a reference audio snippet of a target speaker, the task is to identify the temporal intervals in which the target speaker is present or absent. By incorporating both active speech detection and absence verification, this task evaluates target-voice activity detection.

\paragraph{Voice Change Detection (VCD)} This task requires the model to identify speaker transition points. For non-overlapping speech samples, the focus is on detecting turn-taking boundaries, whereas for overlapping samples, the task involves detecting time intervals containing simultaneous speech.

\subsubsection{Attribution}
This axis evaluates the ability to bind semantic content to its corresponding voice identity in both directions: text to audio and audio to text.

\paragraph{Content-to-Voice Attribution (CVA)} Given a transcribed utterance as a textual query, the model must identify the corresponding speaker from a set of five candidate voices.

\paragraph{Voice-to-Content Attribution (VCA)} In this task, the model is given a brief audio sample of a reference voice and must identify which of five textual candidates was spoken by that voice.

\subsubsection{Reasoning}
\label{sec:reasoning}
This axis examines basic \emph{speaker-attributed reasoning} capabilities. Every Reasoning item is a semantic-hallucination pair: each original clip is paired with a counterfactual variant in which one answer-relevant utterance is re-spoken in another speaker's voice, leaving the transcript unchanged while flipping the correct answer, so that a voice-blind model necessarily fails the pair (Section~\ref{sec:probing_semantic_hallucination}).

\paragraph{Identity Reasoning (IR)} It evaluates whether a model can determine which utterances were spoken by the same person. Given an anchor utterance (text), the model must identify another utterance (text) produced by the same speaker.

\paragraph{Temporal Reasoning (TR)} This task assesses the model's ability to discern sequential relationships and temporal dynamics conditioned on speaker identity. We decompose this evaluation into two granular levels. (a) \emph{Intra-speaker temporal ordering} requires the model to navigate a single speaker's history to spot the first or the last utterance of the speaker. (b) \emph{Inter-speaker temporal alignment} evaluates the ability to align timelines across multiple participants by locating a target speaker's utterance relative to an anchor utterance from a different speaker (e.g., After someone said `X', what was the first thing said by the person who said `Y'?).

\paragraph{Quantitative Reasoning (QR)} This task examines the model's capability to aggregate per-speaker utterance statistics to deduce numerical and comparative relationships. We evaluate this through two subtasks. (a) \emph{Conditional counting} requires the model to quantify the number of distinct speakers who utter a specific word. (b) \emph{Ordinal ranking} assesses the ability to rank speakers by cumulative speaking time and identify the speaker with the $k$-th longest total speaking time (e.g., ``Which of the following was said by the speaker with the $k$-th longest total speaking time?'').

\subsection{Benchmark Construction}
HEAR consists of approximately 2.4K MCQA queries derived from 887 audio clips, totaling roughly 23 hours of audio. The clips average $\sim$93 seconds in length, with a maximum duration of 150 seconds. We curate audio clips exclusively from the test splits of AMI~\cite{10.1007/11677482_3}, ICSI~\cite{Janin2003TheIM}, and VoxMM~\cite{10446300}, where speaker diarization and transcripts are already annotated by humans, ensuring reliable speaker-speech alignments across various real-world acoustic conditions such as meetings, daily conversations, broadcast media, and lectures. Each clip contains 4.05 speakers on average, and all samples are annotated by human annotators as overlap or non-overlap, based on whether answering the question requires focusing on regions with overlapping speech (Appendix~\ref{app:overlap_labeling_Criteria}). We generate queries through a rule-based pipeline and conduct an additional manual verification with eight annotators. As a result, every item in HEAR benefits from two stages of human validation: first during the construction of the source datasets, and again through our additional verification process, yielding a high-quality dataset with strong annotation reliability.

\begin{figure*}[ht]
    \centering
    \includegraphics[width=\linewidth]{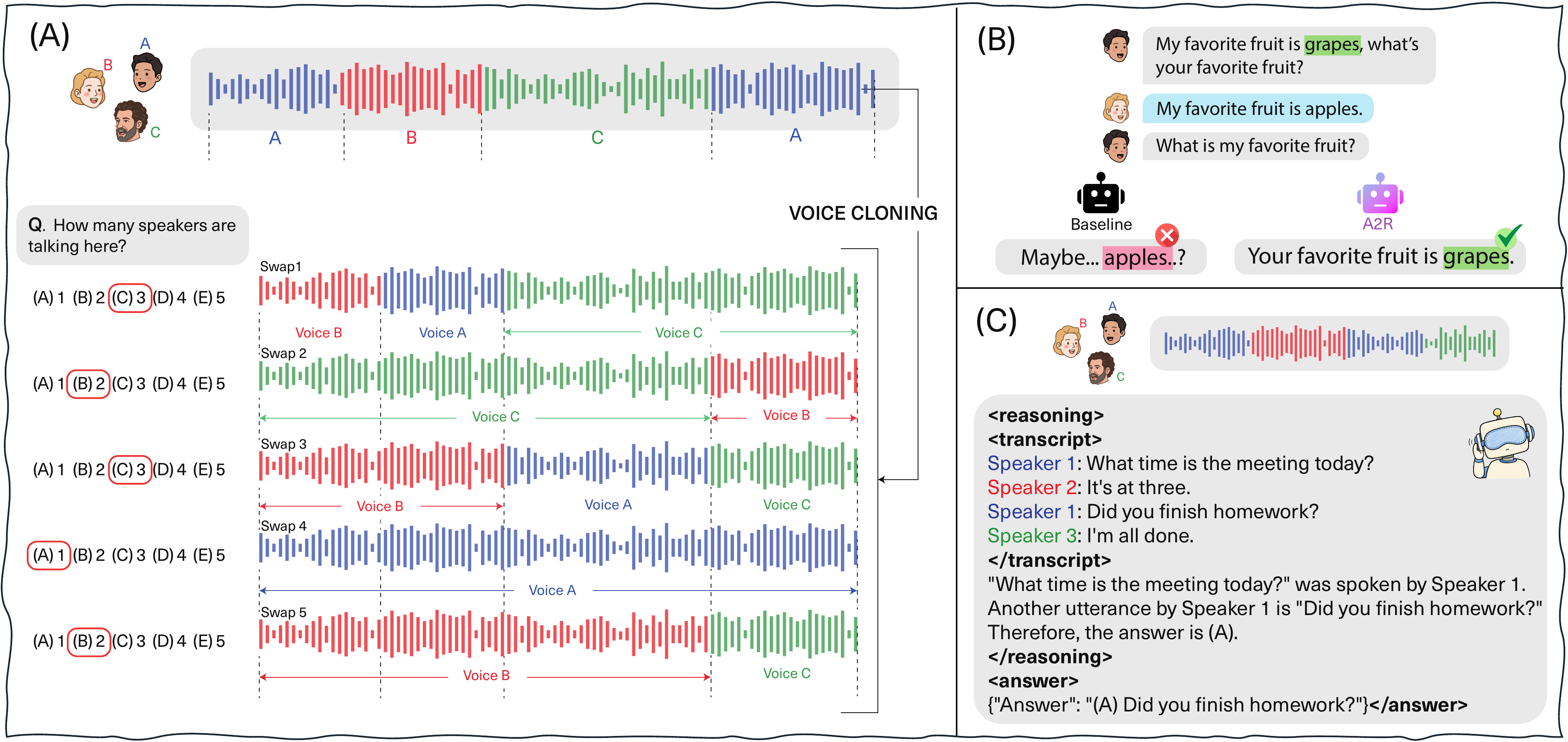}
    \caption{\textbf{Overview of our dataset and method on speaker-attributed reasoning.} \textbf{(A)} Our dataset, CASH, randomly reassigns utterances to different speakers via voice cloning while preserving the whole utterance, yielding counterfactual variants whose ground-truth answer always flips. \textbf{(B)} We show an elementary speaker-attribution task from the \textit{What Do You Like?} benchmark~\cite{wu2024justasrllm}, where the query is unsolvable without identifying which speaker said what. \textbf{(C)} Our model first transcribes the multi-party audio scene, then reasons over the transcript for more reliable speaker-aware reasoning.}
    \label{fig:voice_swap}
\end{figure*}

\section{Training}
\label{training}
\subsection{Motivation}
Unimodal collapse, where a model over-relies on a single modality, has long been a recognized issue in vision-language tasks~\cite{koishigarina2026clipbehaveslikebagofwords}, and a similar phenomenon is observed in the speech domain. A recent study~\cite{lee2026usevaluatingimprovingvoice} highlights that SLMs often exhibit \emph{semantic hallucinations}, bypassing acoustic cues in favor of their textual priors, which is a critical bottleneck for speaker-attributed reasoning in multi-party audio scenes. Determining “who said what” based on semantic plausibility may suggest who \emph{might} speak, but fails in dynamic conversations where any participant can make any statement, thus requiring acoustic vocal cues to be prioritized over textual context to determine who \emph{actually} spoke.

\subsection{Forcing Models to Rely on Vocal Cues}
To encourage models to ground speaker attribution in acoustic vocal cues rather than semantic content, we introduce \textit{Counterfactual Audio with Speaker-level Hard negatives} (\textbf{CASH}). CASH is a 60K-scale dataset for multi-party audio, spanning all task dimensions of HEAR. Given an original multi-speaker clip, we construct counterfactual variants by replacing a target utterance with a voice-converted version of the \emph{same transcript} by a different speaker, while preserving chronological structure and other speech content (Figure~\ref{fig:voice_swap}-(A)).
We build on the VoxMM~\cite{10446300} train split, extracting $\sim$5.2K real-world audio clips drawn from diverse audio scenes. All synthetic voices are generated with VoxCPM2~\cite{zhou2026voxcpm2technicalreport}, yielding $\sim$20K \textit{mixed real-synthetic} audio clips and $\sim$60K queries in total. Details are provided in Appendix~\ref{app:curation_cash}.

\subsection{Rewarding Attribution Before Reasoning}
\label{subsec:reward}
We present \textbf{A2R} (Attribution-to-Reasoning), a model developed by optimizing the baseline model via GRPO to enhance speaker attribution capabilities while preserving general performance. By bridging newly acquired vocal grounding ability with its inherent inferential capacity, we unlock the model's ability to perform speaker-attributed reasoning. During training, as illustrated in Figure~\ref{fig:voice_swap}-(C), the model first transcribes the input audio with speaker tags and then generates both the reasoning and the final answer based on its transcript.

We compute the reward from three aspects of the model response: the speaker-tagged transcription ($R_{\mathrm{tr}}$), the final answer ($R_{\mathrm{ans}}$), and the response format ($R_{\mathrm{fmt}}$). The speaker-tagged transcription not only captures how accurately the model attributes utterances to speakers, but also serves as the basis for the reasoning needed to answer the question, whereas the final answer captures direct task-level correctness. Accordingly, for a model response $Y_i$, we parse its predicted speaker-tagged transcription and answer as $\hat{\tau}_i$ and $\hat{a}_i$, and compare them against the ground-truth transcription $\tau^\star$ and answer $a^\star$. The overall reward is then defined as
\[
\small
\begin{aligned}
&R(Y_i;\tau^\star,a^\star) \\
&\;= R_{\mathrm{tr}}(\hat{\tau}_i;\tau^\star)
 + \lambda_{\mathrm{ans}}R_{\mathrm{ans}}(\hat{a}_i;a^\star)
 + \lambda_{\mathrm{fmt}}R_{\mathrm{fmt}}(Y_i), \\
&R_{\mathrm{tr}}(\hat{\tau}_i;\tau^\star) \\
&\;= \lambda_{\mathrm{cp}}R_{\mathrm{cp}}(\hat{\tau}_i;\tau^\star)
 + \lambda_{\mathrm{cnt}}R_{\mathrm{cnt}}(\hat{\tau}_i;\tau^\star)
 + \lambda_{\mathrm{ord}}R_{\mathrm{ord}}(\hat{\tau}_i;\tau^\star)
\end{aligned}
\]

Here, the transcription reward $R_{\mathrm{tr}}$ is further decomposed into three terms: a cpWER-based~\cite{vonneumann2024meetevaltoolkitcomputationword} attribution reward ($R_{\mathrm{cp}}$), which measures whether the model correctly captures what was said by whom; a speaker-count reward ($R_{\mathrm{cnt}}$), which evaluates whether the predicted transcription contains the correct number of speakers; and a speaker-order reward ($R_{\mathrm{ord}}$), which assesses whether the temporal sequence of speaker turns is preserved (Appendix~\ref{app:reward-details}).

During training, we sample \(G\) responses per input \(x\), compute rewards, and normalize them as group-relative advantages \(A_i=(R_i-\mu_R)/(\sigma_R+\delta)\), where \(\mu_R\) and \(\sigma_R\) are the group reward mean and standard deviation. We then optimize the current policy $\pi_\theta$ using the objective:
\[
\normalsize
\begin{aligned}
\mathcal{L}_{\mathrm{GRPO}}
&=
-\frac{1}{G}\sum_{i=1}^{G}
\min\!\left(r_iA_i,\bar{r}_iA_i\right), \\
\bar{r}_i
&=
\mathrm{clip}(r_i,1-\epsilon_{\mathrm{low}},1+\epsilon_{\mathrm{high}}),
\end{aligned}
\]
where \(r_i=\pi_\theta(Y_i\mid x)/\pi_{\theta_{\mathrm{old}}}(Y_i\mid x)\), with $\pi_{\theta_{\mathrm{old}}}$ denoting the old policy.
\begin{table*}[t]
\centering
\small
\vspace{2mm}
\resizebox{\textwidth}{!}{
\begin{tabular}{@{}l c ccc cc ccc c@{}}
\toprule
 & & \multicolumn{3}{c}{\textbf{Discrimination}} & \multicolumn{2}{c}{\textbf{Attribution}} & \multicolumn{3}{c}{\textbf{Reasoning}} & \\
\cmidrule(lr){3-5} \cmidrule(lr){6-7} \cmidrule(lr){8-10}
\textbf{Model} & \textbf{Size} & \textbf{VC} & \textbf{VCD} & \textbf{VL} & \textbf{CVA} & \textbf{VCA} & \textbf{IR} & \textbf{QR} & \textbf{TR} & \textbf{Avg. (\%)} \\
\midrule
\multicolumn{11}{c}{\textbf{\textit{Proprietary Models}}} \\
\midrule
\geminilogo \, Gemini-3.1-Pro-preview & - & \textbf{68.4} & \textbf{89.5} & \textbf{93.2} & \textbf{94.1} & \textbf{86.4} & \textbf{96.2 (47.6)} & \textbf{89.3 (53.6)} & \textbf{91.3 (51.3)} & \textbf{88.8 (75.8)} \\
\geminilogo \, Gemini-3-Flash-preview & - & 52.6 & 66.0 & 81.4 & 73.9 & 60.1 & 91.9 (40.5) & 82.5 (39.3) & 80.9 (38.3) & 73.1 (58.5) \\
\openailogo \, GPT-4o-audio-preview & - & 40.5 & 21.0 & 28.3 & 60.3 & 43.5 & 80.0 (15.7) & 53.6 (15.7) & 69.6 (17.4) & 48.5 (32.8) \\
\midrule
\multicolumn{11}{c}{\textbf{\textit{Open-Source Models}}} \\
\midrule
\qwenlogo \, Qwen3-Omni-A3B-Thinking & 30B & 34.2 & 33.0 & 31.2 & 45.0 & 34.6 & 71.4 (17.3) & 58.2 (27.5) & 65.2 (12.2) & 45.0 (31.4) \\
\stepfunlogo \, Step-Audio-R1 & 32B & 29.5 & 22.5 & 23.6 & 41.0 & 34.6 & 71.4 (5.4) & 66.8 (22.1) & 67.8 (19.1) & 43.2 (26.5) \\
\qwenlogo \, Qwen3-Omni-A3B-Instruct & 30B & 31.1 & 34.0 & 39.2 & 26.1 & 28.9 & 59.5 (8.6) & 28.6 (5.7) & 57.4 (15.7) & 35.4 (24.1) \\
\openbmblogo \, MiniCPM-o-4.5 & 9B & 17.4 & 22.5 & 34.2 & 35.5 & 29.2 & 59.5 (8.6) & 52.5 (15.4) & 47.8 (10.4) & 36.8 (23.5) \\
\qwenlogo \, Fun-Audio-Chat & 8B & 55.3 & 20.0 & 25.3 & 32.2 & 21.9 & 53.0 (5.9) & 31.1 (6.8) & 45.2 (7.8) & 33.4 (22.5) \\
\kimilogo \, Kimi-Audio-Instruct & 7B & 16.8 & 18.0 & 21.1 & 36.8 & 26.9 & 49.2 (8.1) & 32.5 (2.1) & 53.9 (13.9) & 30.6 (19.2) \\
\mistrallogo \, Voxtral-Small & 24B & 20.0 & 19.0 & 30.8 & 26.1 & 22.3 & 46.5 (3.2) & 36.4 (4.3) & 33.9 (0.0) & 28.8 (17.3) \\
\qwenlogo \, Qwen2.5-Omni & 7B & 12.1 & 20.0 & 31.6 & 25.4 & 24.6 & 48.6 (2.7) & 35.4 (1.4) & 34.8 (2.6) & 28.6 (16.6) \\
\mistrallogo \, Voxtral-Mini & 3B & 30.5 & 20.0 & 17.7 & 24.8 & 25.6 & 33.0 (1.1) & 28.6 (1.4) & 34.8 (1.7) & 26.1 (16.6) \\
\glmlogo \, MiDashengLM & 7B & 34.7 & 19.0 & 26.6 & 19.5 & 20.3 & 24.3 (0.5) & 33.2 (4.3) & 26.1 (0.9) & 25.1 (16.6) \\
\stepfunlogo \, Step-Audio 2 mini & 8B & 13.2 & 20.0 & 24.1 & 23.1 & 19.9 & 35.1 (3.2) & 28.9 (6.4) & 21.7 (2.6) & 23.4 (15.4) \\
\audioflamingologo \, Audio-Flamingo-3 & 7B & 30.5 & 19.5 & 19.8 & 11.7 & 21.6 & 28.6 (3.8) & 24.3 (6.1) & 29.6 (2.6) & 22.0 (15.0) \\
\gemmalogo \, Gemma-4-E4B-it & 4B & 11.6 & 19.5 & 24.1 & 22.5 & 20.3 & 45.4 (2.2) & 24.3 (5.4) & 28.7 (3.5) & 23.9 (14.9) \\
\qwenlogo \, Qwen2.5-Omni & 3B & 10.5 & 21.5 & 22.8 & 26.1 & 21.6 & 35.1 (1.6) & 24.3 (1.4) & 33.0 (0.9) & 23.9 (14.9) \\
\microsoftlogo \, Phi-4-MM & 6B & 21.6 & 20.0 & 21.1 & 19.5 & 20.3 & 28.6 (1.1) & 31.1 (2.5) & 30.4 (0.9) & 23.5 (14.4) \\
\openbmblogo \, MiniCPM-o-2.6 & 8B & 8.4 & 16.0 & 21.9 & 14.3 & 18.3 & 52.4 (3.8) & 41.8 (9.6) & 40.9 (2.6) & 25.3 (13.0) \\
\qwenlogo \, Qwen2-Audio & 7B & 7.9 & 18.5 & 14.8 & 15.6 & 23.9 & 31.4 (0.5) & 46.8 (2.5) & 20.0 (0.0) & 23.1 (11.8) \\
\midrule
\textit{Random Choice} & - & \textit{20.0} & \textit{20.0} & \textit{20.0} & \textit{20.0} & \textit{20.0} & \textit{20.0 (4.0)} & \textit{20.0 (4.0)} & \textit{20.0 (4.0)} & \textit{20.0 (14.9)} \\
\midrule
A2R (Ours) & 30B & \textbf{57.5} & \textbf{38.7} & \textbf{48.4} & \textbf{80.1} & \textbf{70.1} & \textbf{91.6 (70.8)} & \textbf{68.9 (47.4)} & \textbf{84.9 (66.9)} & \textbf{67.1 (60.5)} \\
\bottomrule
\end{tabular}
}
\caption{\textbf{Evaluation results of 20 leading Speech Language Models on the HEAR benchmark.} Values in parentheses in the Reasoning columns report scores under the paired setting, where a prediction is counted as correct only when both the original audio clip and its corresponding semantic hallucination counterpart are correct. A2R results are averaged over three seeds.}
\label{tab:hear_results}
\end{table*}

\section{Experiments}
\label{experiments}
We design our experiments to answer the following three questions: 
(i) How do current leading SLMs perform on HEAR, and do they truly exploit acoustic speaker cues rather than relying on semantic priors for speaker-attributed reasoning? 
(ii) Do the resulting gains transfer to other multi-party audio understanding benchmarks where speaker attribution is essential, while preserving performance in dyadic conversational settings?
(iii) How much does CASH-60K improve speaker-attributed reasoning, and what are the contributions of speaker-level hard negatives and training method? 

\paragraph{Baseline Models.} We evaluate 20 leading SLMs, covering both open-source and proprietary models. 
We assess (a) omni-modality models, (b) speech/audio language models, and (c) proprietary models as closed-source baselines. 
The evaluated models and their inference hyperparameters are summarized in Appendix~\ref{appendix:baseline_hparams} and Table~\ref{tab:baseline_hparams}. Among these models, we choose Qwen3-Omni-30B-A3B-Instruct as our base model for developing speaker-attributed reasoning capabilities, as its demonstrated zero-shot capability on HEAR indicates a promising starting point (Table~\ref{tab:hear_results}). We train LoRA adapters~\cite{hu2021loralowrankadaptationlarge} with $r=64$ on NVIDIA H200 GPUs, while freezing the audio encoders and modality aligners. Details are provided in Appendix~\ref{app:training_details}.

\paragraph{Probing Semantic Hallucination.}
\label{sec:probing_semantic_hallucination}
As introduced in Section~\ref{sec:reasoning}, each ``Reasoning'' pair consists of $(x, x')$, where $x'$ is a CASH-style voice-swapped variant of audio clip $x$. Pair construction enforces five constraints simultaneously: (1) the transcript and temporal structure of $x'$ are preserved relative to $x$; (2) the speaker similarity~\cite{Desplanques_2020} between the swapped target utterance and its reference voice exceeds $0.7$; (3) the WER between the Whisper-large-v3~\cite{radford2022robustspeechrecognitionlargescale} transcription of the synthesized utterance and the reference transcript is $0$, ensuring linguistic content is preserved; (4) the answer is required to be different between $x$ and $x'$, so that any model that ignores acoustic identity would necessarily answer identically on both, thereby failing the question; and (5) $(x, x')$ pairs are retained only if human annotators verify that the synthesized utterance matches the intended target speaker. Target utterances are synthesized with IndexTTS 2~\cite{zhou2025indextts2breakthroughemotionallyexpressive}, a different synthesizer from the one employed in CASH to avoid synthesizer-specific artifacts. Each pair is independently checked by expert annotators for acoustic naturalness, transcript fidelity, and answer-flip validity; pairs failing any criterion are discarded (Appendix~\ref{app:curation_hear}).

\paragraph{Zero-Shot Transfer and Dyadic Evaluation.}
\label{sec:zeroshot_transfer_dyadic_eval}
To evaluate zero-shot transfer beyond HEAR, we consider three held-out benchmarks that require speaker attribution under diverse task settings.

The \textit{What Do You Like? (WDYL)}~\cite{wu2024justasrllm} tests whether a model can associate spoken content with the correct speaker. We focus on its Identity-Critical Questions (ICQ) category, where two speakers describe their preferences and the model must determine whose preference is queried (Figure~\ref{fig:voice_swap}-(B)). We synthesize the last query sentence using a reference voice identical to one of the speakers. The model must therefore identify the queried speaker from vocal characteristics rather than relying on semantic content alone.

We additionally evaluate on \textit{Gaokao}~\cite{hu2024wavllm}, a multi-turn dialogue QA benchmark derived from English listening examinations for the Chinese college entrance exam. To introduce an explicit speaker-attribution requirement, we synthesize the final query using the voice of a randomly selected character from the preceding dialogue, such that answering correctly requires associating the query voice with that character's earlier utterances. We further filter the examples using GPT-5.5~\cite{openai_gpt55_2026} to retain only questions that cannot be solved without speaker attribution.

\begin{figure*}[h]
    \centering
    \includegraphics[width=1\linewidth]{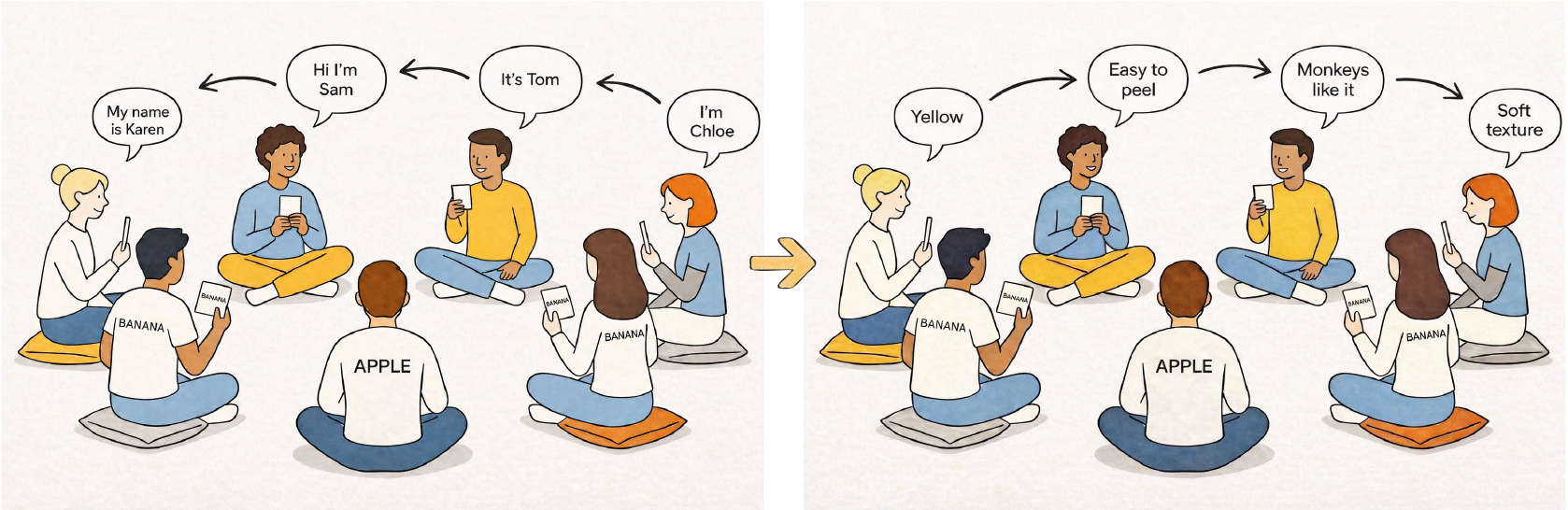}
    \caption{\textbf{Illustration of Find the Spy, a benchmark where speaker attribution is essential.} Each player receives a word (e.g., \textit{Banana}), except for one spy who receives a different word (e.g., \textit{Apple}), and players describe their words in a random speaking order. To identify the spy, a model must do more than recognize which clue is inconsistent with the others: it must also determine \emph{who produced that clue}.}
    \label{fig:FTS_play_figure}
\end{figure*}

We also utilize the \textit{Find the Spy (FTS)}, social deduction game~\cite{xu2025socialmazebenchmarkevaluatingsocial}. 
In this task, 3-6 players introduce themselves and describe their assigned word, while one spy receives a different word. 
The players take turns giving clues about their word, and the group (model) must vote to identify the spy (Figure~\ref{fig:FTS_play_figure}).
Solving the task requires speaker attribution to track which player provided each description, as well as higher-level reasoning to identify the spy whose description is semantically inconsistent with the others. 

Apart from the main audio already provided in the original benchmark, the text query segments are synthesized using a TTS model~\cite{zhou2025indextts2breakthroughemotionallyexpressive}, with speaker similarity constrained to be over 0.7 (Appendix~\ref{app:benchmark_details}).

Lastly, in order to assess whether our model preserves dyadic spoken-language ability, we further evaluate on VoiceBench~\cite{chen2024voicebenchbenchmarkingllmbasedvoice}, which covers eight benchmarks in dyadic conversational settings (Appendix~\ref{app:voicebench}).

\section{Results \& Analysis}

\begin{figure}[!h]
    \centering
    \includegraphics[width=1\linewidth]{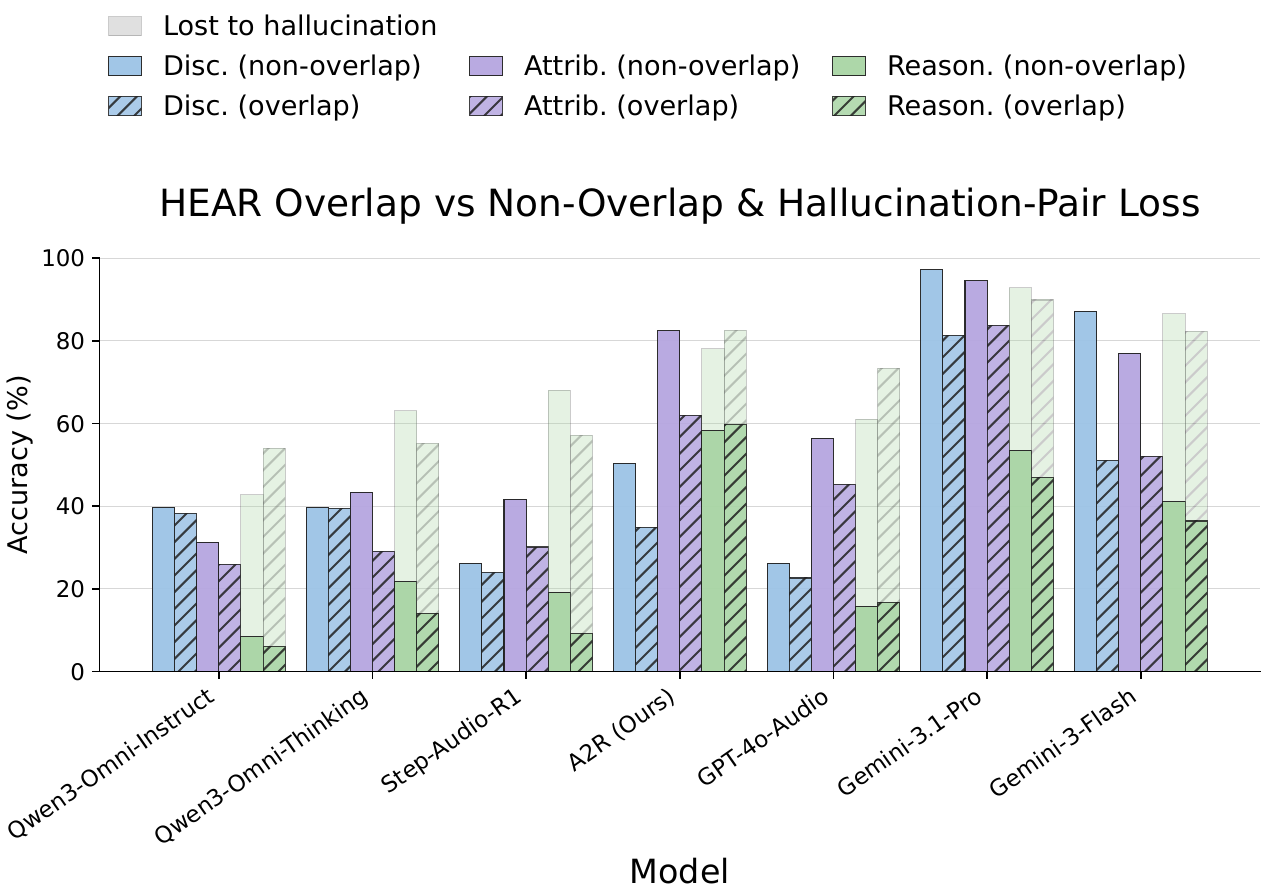}
    \caption{\textbf{Performance degradation under overlap samples and the effect of semantic hallucination.}}
    \label{fig:overlap}
\end{figure}

\paragraph{Limitations of Leading SLMs on HEAR.} Across discrimination and attribution dimensions, most open-source models perform close to random chance, suggesting that current speech-language training does not reliably endow models with the basic capacity to segregate voices and localize them. As illustrated in Figure~\ref{fig:overlap}, this difficulty is exacerbated in scenarios with overlapping speech; performance consistently degrades across six leading baselines. While several recent open-source models achieve modest improvements over chance, the closed-source Gemini series performs substantially better, indicating that the core skills evaluated by HEAR are attainable but remain underdeveloped in current open-source SLMs. Notably, A2R (Ours) largely mitigates this limitation, bringing performance substantially similar to Gemini Flash and significantly narrowing the gap between open- and closed-source models.

\paragraph{Overestimated Reasoning Performance Due to Linguistic Priors.} Open-source models achieve strong reasoning scores, yet their performance drops sharply under the paired accuracy metric reported in parentheses in Table~\ref{tab:hear_results}. The parenthesized score requires a model to answer both an original clip and its semantic-hallucination counterpart correctly (Appendix~\ref{app:paired_reasoning}). The same pattern also appears in proprietary models, including the Gemini models, suggesting that their reasoning also partly relies on semantic cues rather than voice-grounded inference.

\paragraph{Transfers to Unseen Attribution-Critical Tasks.} 
As shown in Table~\ref{tab:ood}, even on elementary tasks such as \textit{What Do You Like?}, the Qwen3 series collapses to near-random performance, reflecting its lack of speaker attribution ability. In contrast, A2R shows substantial gains of +36.4 (WDYL), +22.6 (GAOKAO), and +19.0 (FTS) percentage points. These results suggest that learning speaker attribution provides a transferable inductive bias, enabling the model to generalize speaker-attributed reasoning to unseen tasks without task-specific training.

\begin{table}[h]
\centering
\small
\setlength{\tabcolsep}{4pt}
\renewcommand{\arraystretch}{1.10}
\resizebox{\columnwidth}{!}{%
\begin{tabular}{l c c c}
\toprule
\textbf{Model} & \textbf{WDYL} & \textbf{GAOKAO} & \textbf{FTS} \\
\midrule
Random Choice                     & 50.0  & 50.0  & 23.8 \\
\midrule
Qwen3-Omni-30B-A3B-Instruct       & 61.3  & 65.6  & 29.2 \\
Qwen3-Omni-30B-A3B-Thinking       & 67.1  & 54.8  & 17.2 \\
\midrule
A2R (CASH w/o HN)            & 95.0  & 78.5  & 40.5 \\
A2R (CASH)             & \textbf{97.7}  & \textbf{88.2}  & \textbf{48.2} \\
\bottomrule
\end{tabular}%
}
\caption{\textbf{Generalization across three benchmarks where speaker attribution is essential.} ``HN'' denotes voice-swapped hard negative variants in CASH.}
\label{tab:ood}
\end{table}

\paragraph{Transfers to Real-World Human Voices.}
To examine whether A2R's gains transfer to real-world human voices, we randomly sample 100 conversations used in the unseen-distribution evaluation above (50 WDYL, 30 GAOKAO, and 20 FTS) and collect corresponding human recordings for each instance. Specifically, five participants recorded the same conversational transcripts underlying the original synthesized inputs, with speakers randomly assigned to conversational roles. As shown in Table~\ref{tab:human_ood}, A2R's substantial improvements over the base model persist when synthesized speech is replaced by human recordings: on WDYL, GAOKAO, and FTS, A2R improves over the base model by +48.0, +20.0, and +30.0 percentage points, respectively. These results provide evidence that the attribution capability learned from CASH transfers to real-world human voices beyond synthesized speech.

\begin{table}[!h]
\centering
\small
\setlength{\tabcolsep}{4pt}
\renewcommand{\arraystretch}{1.10}
\resizebox{\columnwidth}{!}{%
\begin{tabular}{l c c c c}
\toprule
\textbf{Model} & \textbf{Audio} & \textbf{WDYL} & \textbf{GAOKAO} & \textbf{FTS} \\
\midrule
Random Choice                  & -   & 50.0  & 50.0  & 23.8 \\
\midrule
Qwen3-Omni-30B-A3B-Instruct
    & Synthetic & 50.0 & 76.7 & 50.0 \\
A2R (CASH)
    & Synthetic & \textbf{96.0} & \textbf{93.3} & \textbf{65.0} \\
\midrule
Qwen3-Omni-30B-A3B-Instruct
    & Human & 50.0 & 63.3 & 40.0 \\
A2R (CASH)
    & Human & \textbf{98.0} & \textbf{83.3} & \textbf{70.0} \\
\bottomrule
\end{tabular}%
}
\caption{
\textbf{Evaluation on human-recording attribution-critical tasks.}
We compare the 100 synthetic samples with corresponding recordings produced by human speakers.
}
\label{tab:human_ood}
\end{table}

\begin{figure}[!h]
    \centering
    \includegraphics[width=1\linewidth]{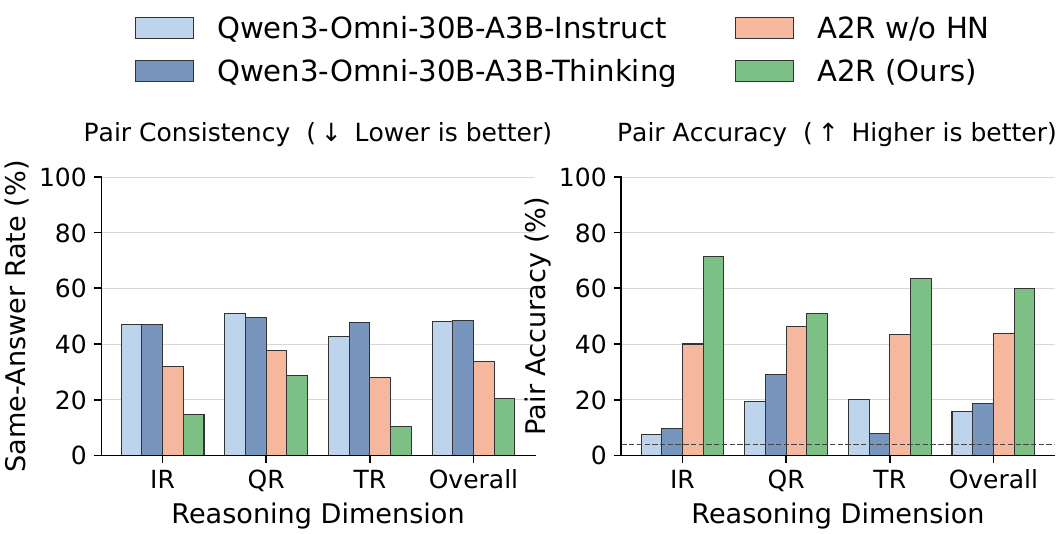}
    \caption{\textbf{Results on semantic hallucination pairs in the ``Reasoning'' category of HEAR.}
        ``HN'' denotes voice-swapped hard negative variants in CASH.
        Lower same-answer rates indicate higher acoustic sensitivity, as each pair is designed to elicit different answers.
        }
    \label{fig:hear_ablation}
\end{figure}

\begin{figure}[!h]
    \centering
    \includegraphics[width=1\linewidth]{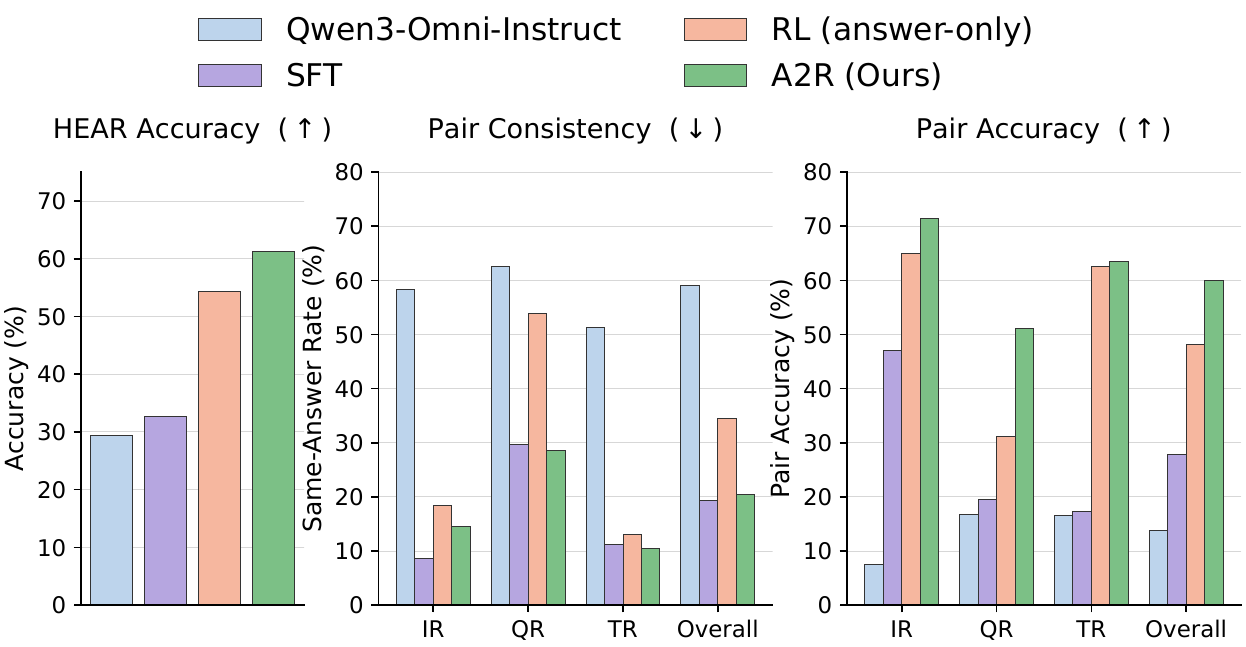}
    \caption{\textbf{Ablation of training regimes on HEAR.}        
        }
    \label{fig:train_ablation}
\end{figure}

\paragraph{Ablation Study.}
To isolate the contribution of each component, we conduct ablations along three axes: (i) the inclusion of speaker-level hard negatives in CASH; (ii) reward shaping, where the model is prompted to reason but is not required to follow our structured reasoning trace, with rewards assigned only to the final answer; and (iii) the training regime, where the baseline model is trained with LoRA-based supervised fine-tuning (SFT). As shown in Figure~\ref{fig:hear_ablation}, removing the hard negatives from CASH substantially reduces HEAR reasoning pair accuracy from 60.0 to 43.0 (for one seed) and also leads to a higher same-answer rate. These results suggest that the HEAR-axis data itself promotes acoustic sensitivity, while the hard-negative structure further strengthens this capability by encouraging the model to rely on speaker-specific voice cues. 
Importantly, the benefits of this structure extend beyond HEAR: as shown in Table~\ref{tab:ood}, incorporating hard negatives also improves performance on zero-shot benchmarks. For the second axis, removing the structured reasoning-trace constraint and rewarding only the final answer similarly degrades overall HEAR performance and pair consistency, as shown in Figure~\ref{fig:train_ablation}. Finally, replacing our training objective with LoRA-based SFT yields weaker results not only on HEAR but also across most VoiceBench settings, as shown in Table~\ref{tab:voicebench}. All ablation models are trained for one epoch and evaluated after validation performance has stabilized.

\paragraph{Minimal Trade-off on General Dyadic Tasks.} As shown in Table~\ref{tab:voicebench}, A2R achieves performance broadly on par with the baseline across eight general dyadic benchmarks, demonstrating that speaker-attributed reasoning can be acquired with only a marginal impact on general capability.

\begin{table}[!h]
\centering
\resizebox{\columnwidth}{!}{%
    \begin{tabular}{l c c } 
    \toprule
    \textbf{Datasets} & \textbf{Model} & \textbf{Performance}
    \\\midrule 
    \multirow{3}{*}{\begin{tabular}[c]{@{}l@{}} 
     \textit{AlpacaEval} \textbar \  \textit{BBH} \textbar \\ 
     \textit{AdvBench} \textbar \ \textit{CommonEval} 
    \end{tabular}}  
    
     & Baseline             & 4.38 $|$ 94.00 $|$ 99.42 $|$ 4.07
    \\ & A2R                & 4.31 $|$ 93.90 $|$ 96.35 $|$ 3.75
    \\ & Ablation (SFT)     & 3.19 $|$ 68.10 $|$ 89.62 $|$ 1.90
    
    \\ \midrule 
    \multirow{3}{*}{\begin{tabular}[c]{@{}l@{}} 
     \textit{WildVoice} \textbar \ \textit{IFEval} \textbar \\ 
     \textit{MMSU} \textbar \ \textit{OpenBookQA} 
    \end{tabular}}
           
     & Baseline            & 4.15 $|$ 87.97 $|$ 80.84 $|$ 95.38
    \\ & A2R               & 3.93 $|$ 86.01 $|$ 76.87 $|$ 92.53
    \\ & Ablation (SFT)    & 2.14 $|$ 78.05 $|$ 75.73 $|$ 90.33
    \\
    \bottomrule
    \end{tabular}%
}
\caption{\textbf{Evaluation on VoiceBench.} We assess performance drop under dyadic conversational settings.}
\label{tab:voicebench}
\end{table}

\begin{figure}[!h]
    \centering
    \includegraphics[width=\linewidth]{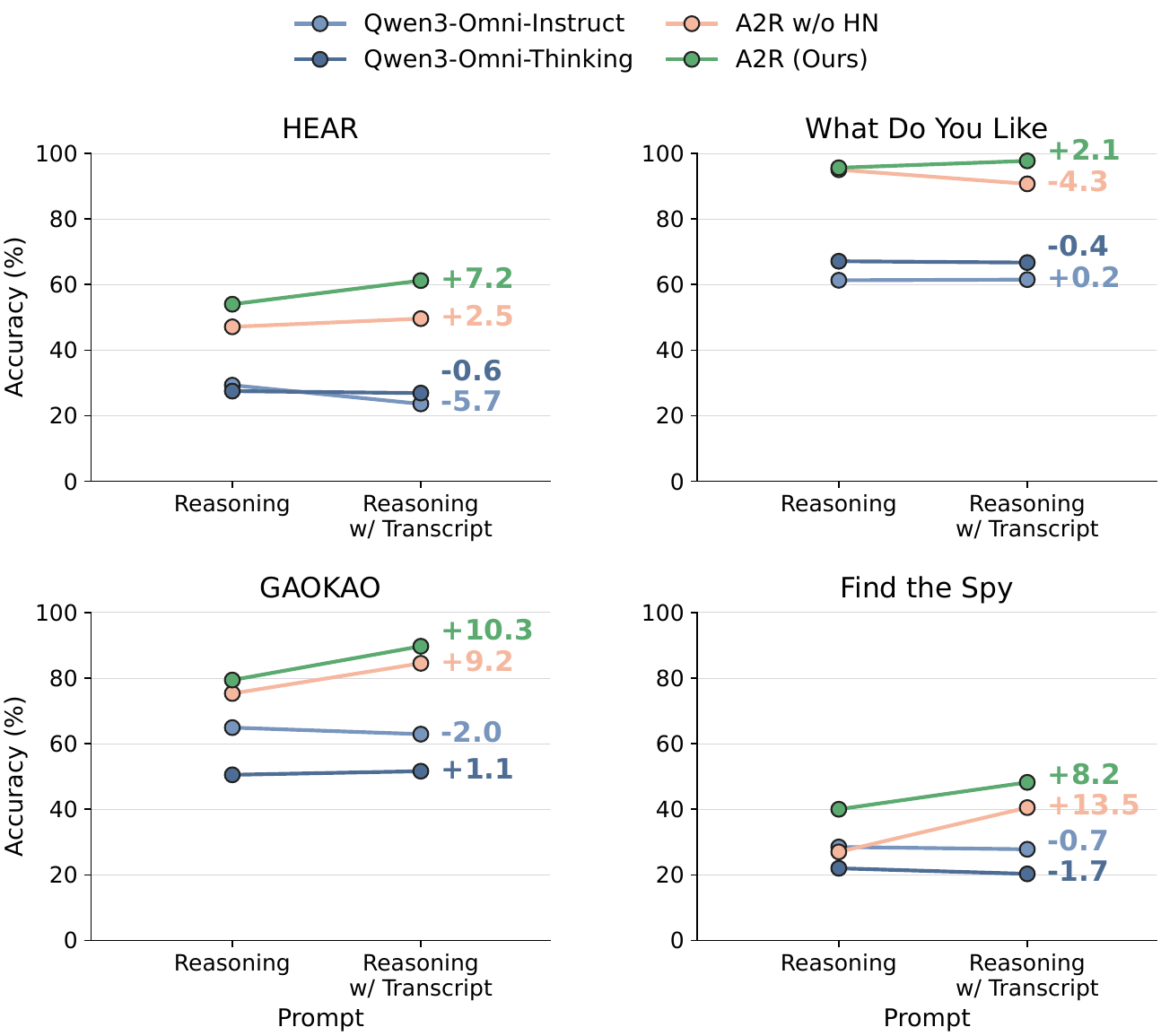}
    \caption{\textbf{Prompt Analysis on benchmarks.} ``Reasoning'' denotes a prompt that asks the model to provide brief reasoning before answering, while ``Reasoning w/ transcript'' asks for a transcript before answering.}
    \label{fig:prompt_analysis}
\end{figure}

\paragraph{Prompt Analysis.}
To assess A2R's learned speaker attribution capability independent of transcript-based reasoning, we remove the reasoning trace at inference time. Figure~\ref{fig:prompt_analysis} compares \textit{Reasoning w/ Transcript}, which explicitly generates a transcript before reasoning, with a less structured \textit{Reasoning} prompt that elicits only a brief rationale. While transcript-first reasoning provides an additional benefit, A2R continues to outperform the baselines on the majority of benchmarks even without this scaffold. This suggests that A2R has internalized speaker attribution rather than merely learning to exploit a particular prompting strategy.

\paragraph{Why A2R Works.}
In our answer-only ablation experiments, we observed the model relying on subjective adjectives (e.g., ``high pitch'', ``documentary-like voice'') to track speakers. This mapping from continuous acoustic signals to descriptive text can induce identity hallucination. While merely prompting baselines to use structural identifiers (e.g., ``Speaker 1:'') does not help (Figure~\ref{fig:prompt_analysis}), explicitly reinforcing a 1:1 alignment between distinct voices and these tags establishes a reliable anchor for the model's reasoning process. Replacing vague vocal descriptors with discrete tags prevents identity confusion throughout extended chain-of-thought generation.

\section{Related Works}
\paragraph{Speaker Attribution: From Speaker Recognition to Reasoning.}
The question of \emph{who said what}~\cite{kanda2021comparativestudymodularjoint} is traditionally addressed through speaker diarization and multi-talker ASR, which identify \emph{who spoke when} and \emph{what was said}, respectively. Recent LLM-based recognizers improve this pipeline by incorporating speaker information or supporting instruction-guided transcription~\cite{wang2024metacatspeakerinformedspeechembeddings, li2026dmasrdiarizationawaremultispeakerasr, meng2025largelanguagemodeltranscribe}. Yet these systems mainly produce speaker-labeled transcripts, leaving speaker-content reasoning largely unexplored.
We refer to this capability as \emph{speaker-attributed reasoning}: answering questions that require linking what was said to who said it, such as identifying the speaker of an utterance or determining who spoke more or first.

\paragraph{Evaluation of Speaker-Attributed Understanding.}
Recent SLM benchmarks have expanded from single-speaker QA to broader audio and multi-party understanding, but speaker attribution remains only partially isolated. Broad benchmarks such as MMAU~\cite{sakshi2024mmaumassivemultitaskaudio, kumar2025mmauprochallengingcomprehensivebenchmark}, AudioMarathon~\cite{he2025audiomarathoncomprehensivebenchmarklongcontext}, and ChronosAudio~\cite{luo2026chronosaudiocomprehensivelongaudiobenchmark} include multi-speaker scenarios, yet treat them as a small subset of tasks (e.g., Voice Counting) among many evaluation axes. M3-SLU~\cite{kwon2025m3sluevaluatingspeakerattributedreasoning} is closest to our setting, but its evaluation is limited to a narrow task (e.g., similar to IR dimension of HEAR), mainly true/false or single utterance--speaker matching. MSU-Bench~\cite{wang2025msubenchunderstandingconversationalmultitalker} evaluates multi-party conversational understanding, but does not separate speaker attribution as a dedicated axis. In contrast, HEAR isolates speaker attribution itself as the primary axis, organizing it into discrimination, attribution, and reasoning stages that prior benchmarks leave entangled.
\section{Conclusion}
We introduced HEAR, a benchmark for evaluating speaker-attributed reasoning in SLMs, and showed that current models often fail to bind utterances to distinct voices, instead relying on semantic priors over vocal evidence. To address this limitation, we proposed CASH, a 60K-scale dataset of voice-cloned hard negatives that decouples acoustic cues from semantic content, and developed A2R to explicitly learn speaker attribution. A2R substantially improves performance on HEAR and transfers to unseen multi-speaker tasks. Together, we hope our findings establish speaker attribution as a key step toward multi-party auditory comprehension.

\section*{Limitations}
While explicit reasoning via transcript generation maximizes speaker attribution performance, it introduces latency that limits real-time applications. Importantly, externalizing transcripts is not essential: as shown in Figure~\ref{fig:prompt_analysis}, A2R significantly outperforms baselines even without this step, albeit with slightly lower performance. Bridging this latency–performance gap through implicit reasoning over internal representations, without generating explicit reasoning traces, is a promising direction for future work.

\section*{Ethics Statement}
All verification in our work was conducted by eight authors of this paper. Our work also involves voice cloning samples that pose risks of impersonation and speaker re-identification. Because dataset or copyright licenses do not themselves constitute consent for voice cloning, and no specific consent was obtained from source speakers, synthetic audio is restricted to approved, non-commercial research under safeguards against misuse.

\paragraph{Dual-use risks.} Given the potential misuse of voice cloning for impersonation, deceptive audio generation, fraud, harassment, speaker re-identification, and other harmful applications, we treat the synthetic speech component of our work as a controlled research resource rather than an unrestricted public dataset.

\paragraph{Voice, likeness, and consent.} Although the source corpora provide licenses or research-use conditions governing the underlying recordings, these conditions do not necessarily constitute consent to generate novel utterances in an individual’s voice, and we did not obtain additional task-specific consent from the source speakers. We also do not regard pseudonymous speaker identifiers as providing complete anonymity, because a recognizable voice can itself convey identity.

\paragraph{Release scope and access control.} Code, evaluation protocols, non-identifying annotations and metadata, and aggregate statistics can be released publicly, whereas synthetic waveform data are available only through a gated research repository for approved, non-commercial research related to speaker-aware speech understanding and reasoning. Requesters must provide their identity, institutional affiliation, and intended research use and agree to the Data Use Agreement (DUA) before access is granted.

\paragraph{Data Use Agreement.} The DUA restricts use of the synthetic audio to approved non-commercial research and prohibits redistribution, attempts to re-identify source speakers, impersonation, deceptive or misleading use, commercial voice replication, deployment of cloned voices in interactive or production systems, and other uses intended to harm or misrepresent individuals. Access is revocable when these conditions are violated.

\paragraph{Non-attribution and documentation.} All synthetic utterances are documented as artificially generated samples. They are not statements actually made by the corresponding source speakers and must not be interpreted as representing those individuals’ views, intentions, beliefs, or endorsements. Personally identifying metadata that are unnecessary for the research purpose, including available speaker names and source-speaker mappings, are removed from the released resource.

\paragraph{Takedown procedure.} Both rights holders and individuals whose voices occur in the resource, or their authorized representatives, can request removal of corresponding entries. Requests can identify the relevant source recording or other information sufficient to locate the affected samples. Valid requests result in removal of the associated source-derived and synthetic entries from subsequent distributions of the resource. Takedown records are maintained in the dataset documentation without disclosing unnecessary personal information about the requester.

\section*{Acknowledgements}
This work was supported by Institute of Information \& Communications Technology Planning \& Evaluation (IITP) grants funded by the Korea government (MSIT) [NO.RS-2021-II211343, Artificial Intelligence Graduate School Program (Seoul National University); No.2022-0-00959, RS-2022-II220959], National Research Foundation of Korea (NRF) grant [No.2022R1A3B1077720, 2022R1A5A7083908], BK21 FOUR Program of the Education and Research Program for Future ICT Pioneers, Seoul National University in 2026, Mobile eXperience(MX) Business, Samsung Electronics Co., Ltd., NVIDIA Academic Grant Program and the Research Grant from Seoul National University(800-20250397).

\bibliography{custom}

\appendix
\newpage

\appendix

\label{samples}

\section{Curation of CASH \& Statistics}
\label{app:curation_cash}
\paragraph{CASH-60K.} We build on the VoxMM~\cite{10446300} train split, extracting 5{,}242 multi-speaker clips drawn from diverse multi-speaker scenarios such as daily conversations, interviews and broadcast scenes. From these originals we synthesize 14{,}180 voice-swap variant clips, about 2.7 variants per original on average, yielding 19{,}422 audio clips in total. We then use both originals and variants to instantiate the task dimensions defined in Section~\ref{benchmark}, producing 59{,}762 training queries (roughly three queries per audio clip): 31{,}045 for attribution, 20{,}094 for reasoning, and 8{,}623 for discrimination. All voice cloning is performed with VoxCPM2~\cite{zhou2026voxcpm2technicalreport}, one of the current state-of-the-art TTS systems. To ensure the linguistic and acoustic fidelity of the cloned speech, we apply a two-stage filtering protocol to every synthesized clip. First, the clip is transcribed with Whisper-large-v3~\cite{radford2022robustspeechrecognitionlargescale} and we retain only samples whose word error rate (WER) against the target transcript falls below 0.1. Second, we compute the cosine similarity between ECAPA-TDNN~\cite{Desplanques_2020} embeddings of the synthesized clip and the reference voice, and keep only samples whose speaker similarity exceeds 0.7.
\section{Curation Process of HEAR Benchmark}
\label{app:curation_hear}
This section details the data curation methodology for the HEAR benchmark. The curation pipeline operates in two primary stages: an automated process to generate a comprehensive pool of candidate questions from three established datasets, followed by a rigorous manual review phase wherein human annotators validate and filter the questions to ensure high data quality.

\subsection{Rule-Based Question Construction}
\label{app:rule_based}

\paragraph{Source Data and Extraction.}
We leverage three publicly available datasets characterized by human-verified text transcripts: AMI and ICSI (multi-party meetings) and VoxMM (CC-BY-4.0-licensed YouTube videos). To maintain ground-truth fidelity, we strictly rely on the original human annotations, explicitly excluding any AI-generated transcriptions. The source audio is segmented into 30- to 150-second clips (averaging approximately $93$ seconds), with each segment curated to contain at least two distinct speakers, $4.05$ on average and ranging from $2$ to $12$.

\paragraph{Voice References.}
All voice samples utilized as multiple-choice options are sourced externally to the test clip. These external reference samples consistently feature a single, non-overlapping speaker and range in duration from 3 to 10 seconds.

\subsection{Taxonomy}
The automated pipeline generated three categories of questions:
\paragraph{Discrimination.}
The Discrimination dimension is generated mechanically from the transcript and speaker timing.  \textsc{VC} (\emph{voice cardinality}) asks for the number of distinct speakers in a clip.  \textsc{VL} (\emph{voice localization}) presents a reference voice and asks the model to select a time range on the basis of that speaker's activity. Each item takes one of two polarities: a presence variant, whose gold option is an interval in which the target speaker is active while the four distractors are intervals in which the speaker is not, and an absence variant, which reverses this so that the gold option is the only interval containing no speech from the target.  \textsc{VCD} (\emph{voice change/overlap detection}) asks for the time range that contains either a turn change or an overlap; the generator enumerates all eligible 3--7\,s windows and rejects any that
violate the regime constraint. 

\paragraph{Attribution.}
Attribution items are produced by jointly sampling a 5-way option set on top of the per-clip speaker pool.  \textsc{CVA} (\emph{content$\to$voice}) shows a transcript span and asks which voice uttered it; \textsc{VCA} (\emph{voice$\to$content}) shows a voice and asks which transcript span matches it.  Both come in non-overlap and overlap variants determined by the surrounding speech context.  Distractors are drawn from the same clip
to enforce voice-level discrimination; each option's voice reference is the external segment described above.

\paragraph{Reasoning.}
Reasoning items follow a hallucination-pair design.  For every original question, we synthesize a counterfactual variant in which a single answer-relevant utterance has been re-spoken in a \emph{different} speaker's voice via voice cloning, while leaving all other audio untouched.  Voice swaps are produced by IndexTTS 2~\cite{zhou2025indextts2breakthroughemotionallyexpressive} conditioned on external 3--10\,s voice references from the long source clip, and only swaps
whose Whisper-large-v3~\cite{radford2022robustspeechrecognitionlargescale} ASR transcript matches the target text with ${\rm WER}\!=\!0$ are admitted.  Three sub-types are produced: \textsc{QR} (Quantitative Reasoning), \textsc{IR}
(Identity Reasoning), and \textsc{TR} (Temporal Reasoning). A model that ignores voice identity must answer the original and the hallucinated variant identically and therefore fails the pair; this is the mechanism by which the Reasoning split exposes voice-blind shortcutting.

\subsection{Overlap Labeling Criteria}
\label{app:overlap_labeling_Criteria}

To further evaluate robustness under acoustically mixed conditions, we provide overlap-labeled subsets for all taxonomy dimensions. Overlap samples are annotated by human evaluators based on whether the overlapping speech segment is required to derive the correct answer, along with additional criteria specific to each dimension. Below, we describe the overlap criteria and task-specific objectives for each category.

\paragraph{Voice Localization (VL-Overlap)}
The overlap subset is defined as samples where the target speaker's active interval temporally overlaps with speech from at least one additional speaker. The objective is to localize the target voice despite acoustic interference and concurrent speech activity.

\paragraph{Voice Change Detection (VCD-Overlap)}
For overlap samples, the target interval contains simultaneous speech activity from multiple speakers. Unlike the standard setting, which focuses on clean turn-taking boundaries, this subset evaluates whether the model can detect overlap onset and offset boundaries within acoustically mixed regions.

\paragraph{Content-to-Voice Attribution (CVA-Overlap)}
The overlap subset consists of utterances spoken during intervals containing concurrent speech from multiple speakers. The model must correctly associate the queried textual utterance with its corresponding speaker identity despite acoustic mixtures.

\paragraph{Voice-to-Content Attribution (VCA-Overlap)}
The overlap subset is defined by cases where the ground-truth utterance occurs within overlapping speech intervals. The task evaluates whether the model can correctly retrieve the linguistic content associated with a target voice under concurrent speech conditions.

\paragraph{Identity Reasoning (IR-Overlap)}
IR-Overlap samples contain anchor or candidate utterances that temporally intersect with overlapping speech regions. The model must resolve speaker identity consistently despite acoustic interference.

\paragraph{Temporal Reasoning (TR-Overlap)}
The overlap subset includes temporal reasoning chains involving utterances occurring during concurrent speech. Models must correctly infer speaker-conditioned temporal relationships while disentangling overlapping voices.

\paragraph{Quantitative Reasoning (QR-Overlap)}
QR-Overlap samples involve speaker statistics computed from sessions containing overlapping speech intervals. The model must aggregate speaker-specific evidence robustly despite acoustically mixed observations.

\begin{figure*}
    \centering
    \includegraphics[width=1\linewidth]{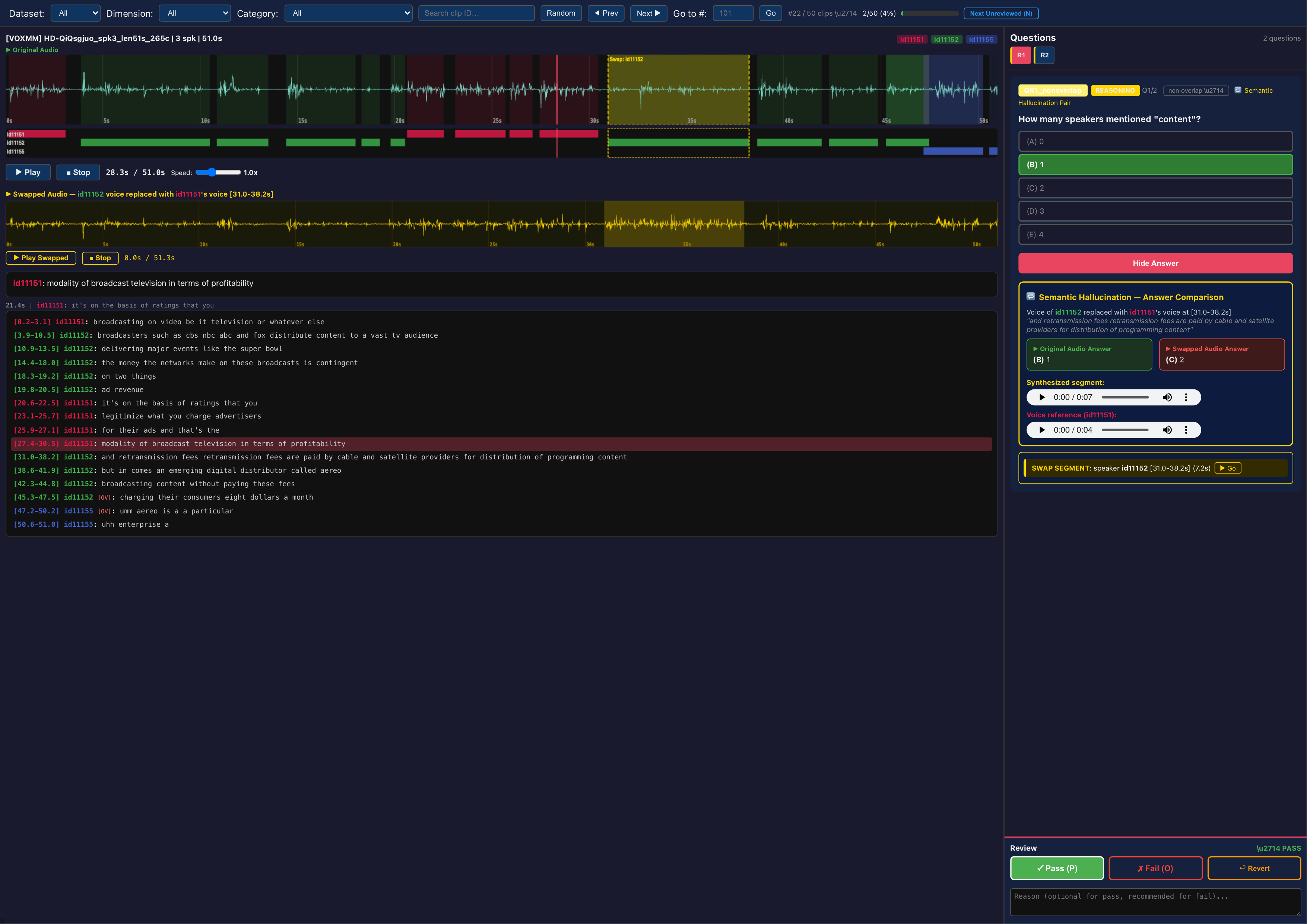}
    \caption{Screenshot of the interface shown to human annotators during verification.}
    \label{fig:human_filtering}
\end{figure*}

\subsection{Human Verification}
\label{app:human_verification}

Eight members of the authors served as annotators. All were graduate students in artificial intelligence and comfortable reading English transcripts and listening to long-form multi-party audio. Annotators worked through a custom local web tool that played the clip audio, displayed the word-level transcript, exposed every option (including each voice reference), and offered Pass/Fail buttons together with a free-text reason field as described in Figure~\ref{fig:human_filtering}.  

\paragraph{Filtering criteria.}
\begin{enumerate}
\item \textbf{Solvability.} The reviewers are requested to answer the question correctly using only the clip audio, transcript, and the offered voice references. Items that are ambiguous, under-specified, or require information outside the clip are rejected. \item \textbf{Unique correct answer.} Exactly one option must be correct and the remaining four distractors must each be falsifiable from the clip. Items with multiple defensible answers, with options that are mutual paraphrases, or whose distractors collapse onto the gold answer are rejected.

\item \textbf{Voice-identity preservation (Reasoning split only).} For each TTS-synthesized swap, the reviewer plays both the original and the swapped clip and confirms that the swapped voice is perceptually \emph{the same person} as the conditioning voice reference, and that the synthesis is free of obvious artifacts (mispronunciation, prosodic discontinuity at the splice, audible bandwidth change).  Pairs that fail either criterion are rejected even if WER\,$=\!0$. \item \textbf{Transcript--option consistency.} Although AMI, ICSI, and VoxMM are all distributed with human-produced transcripts, boundary errors (off-by-a-word starts/ends, missed back-channels, mis-attributed overlapping speech) still survive into our candidate pool.
Any item whose option text or query text fails to match the audio at the specified timestamps, even by a single word, is filtered out. 
\end{enumerate}

After the verification pass, $887$ clips and $2{,}395$ questions survive from a candidate pool of roughly three times that size. The resulting benchmark therefore reflects two compounded layers of human quality control: the original transcripts of the three source corpora, and the per-question audit described above.

\subsection{Benchmark Statistics}
\label{app:stats}

\begin{table}[t]
\centering
\small
\setlength{\tabcolsep}{4pt}
\renewcommand{\arraystretch}{1.05}
\resizebox{\columnwidth}{!}{%
\begin{tabular}{lrrrrrr}
\toprule
\textbf{Domain} & \textbf{\#Clip} & \textbf{\#Src.} &
\textbf{\#Spk.} & \textbf{Dur.} & \textbf{Spk/clip} & \textbf{Hours} \\
\midrule
VoxMM  & 360 & 34 & 245 & 87.8 & 3.64 & 8.78 \\
AMI    & 241 & 23  &  24 & 97.4 & 3.93 & 6.52 \\
ICSI   & 286 &  6   &  21 & 94.9 & 4.65 & 7.54 \\
\midrule
\textbf{TOTAL}  & \textbf{887} & \textbf{63}  &
\textbf{290} & \textbf{92.7} & \textbf{4.05} & \textbf{22.84} \\
\bottomrule
\end{tabular}
}
\caption{HEAR per-domain composition. \textbf{\#Src.}: distinct source recordings.
\label{tab:hear_v2_summary}\textbf{\#Spk.}: speakers, deduplicated within each corpus's identity namespace. \textbf{Dur.}: average clip duration, in seconds. \textbf{Hours}: total speech-clip-hours.}
\end{table}

Table~\ref{tab:hear_v2_summary} reports the per-domain composition of HEAR. The benchmark covers $887$ clips, totals $\sim$\,$22.8$\,h of audio, and exposes a model to $290$ unique speakers in aggregate.  VoxMM contributes the bulk of the speaker diversity ($245$ speakers across $34$ CC-BY-4.0-licensed videos (audio) in $11$ genre categories), and AMI and ICSI contribute longer, denser meeting clips ($3.9$--$4.7$ speakers/clip on average).

\begin{figure*}[t]
\centering
\includegraphics[width=\linewidth]{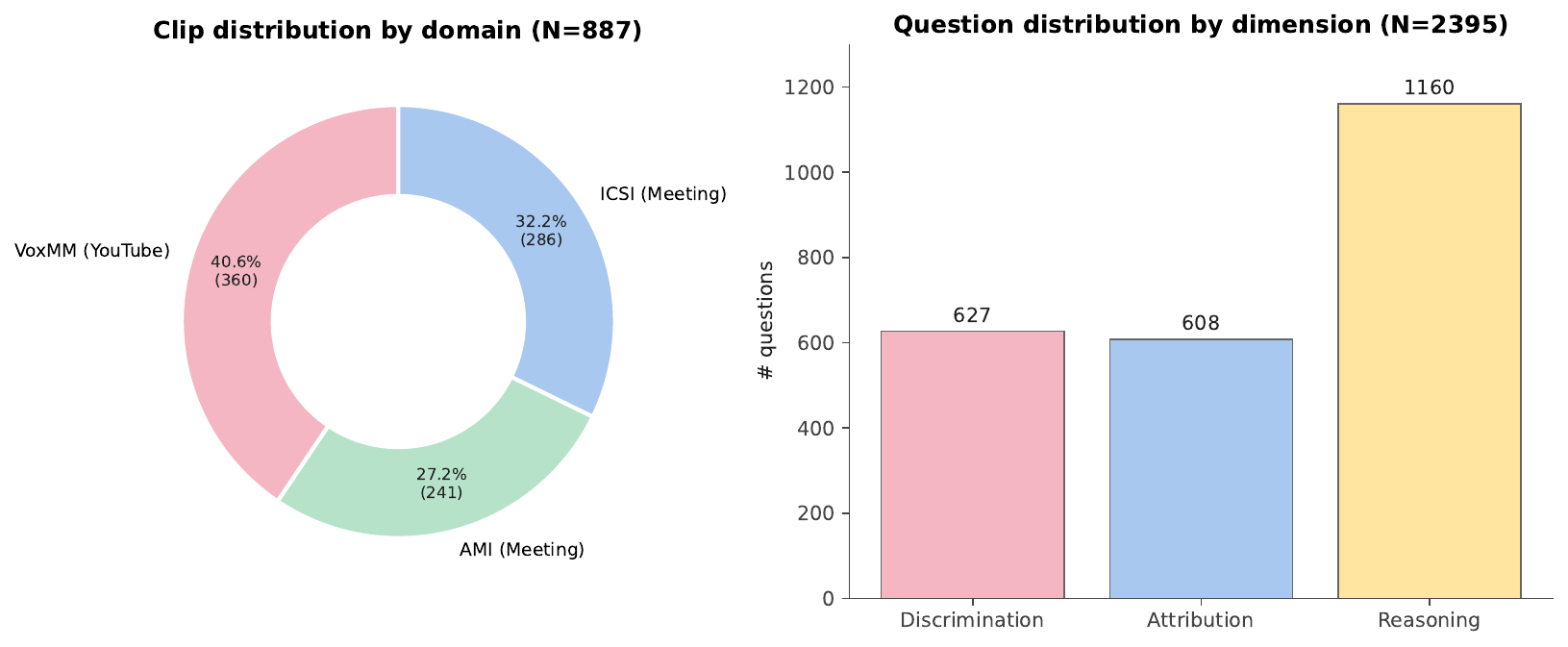}
\caption{HEAR composition.  Left: clip distribution across the three source domains ($N\!=\!887$).  Right: question distribution across the three evaluation dimensions ($N\!=\!2{,}395$).}
\label{fig:domain_dim}
\end{figure*}

\begin{figure*}[t]
\centering
\includegraphics[width=0.85\linewidth]{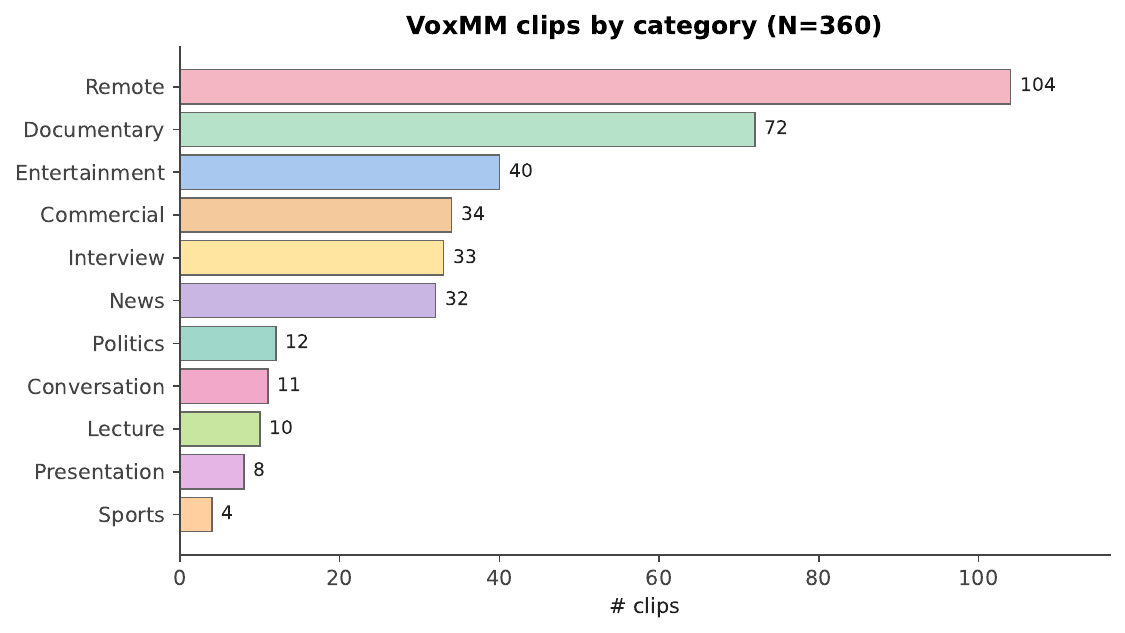}
\caption{Distribution of VoxMM clips in HEAR by genre category
($N\!=\!360$).}
\label{fig:voxmm_categories}
\end{figure*}

\paragraph{Domain and dimension distribution.}
Figure~\ref{fig:domain_dim} (left) shows that no single domain dominates the benchmark: VoxMM accounts for $40.6\%$ of clips, ICSI for $32.2\%$, and AMI for $27.2\%$.  The skew toward meeting-style audio is intentional and follows the original corpus sizes; we counter-balance it on the question side by saturating the VoxMM category budget (Figure~\ref{fig:voxmm_categories}).
Figure~\ref{fig:domain_dim} (right) reports the question count per evaluation dimension: $627$ Discrimination, $608$ Attribution, and $1{,}160$ Reasoning items (original\,$+$\,swapped), for a total of $2{,}395$ questions. 

\begin{figure*}[t]
\centering
\includegraphics[width=0.55\linewidth]{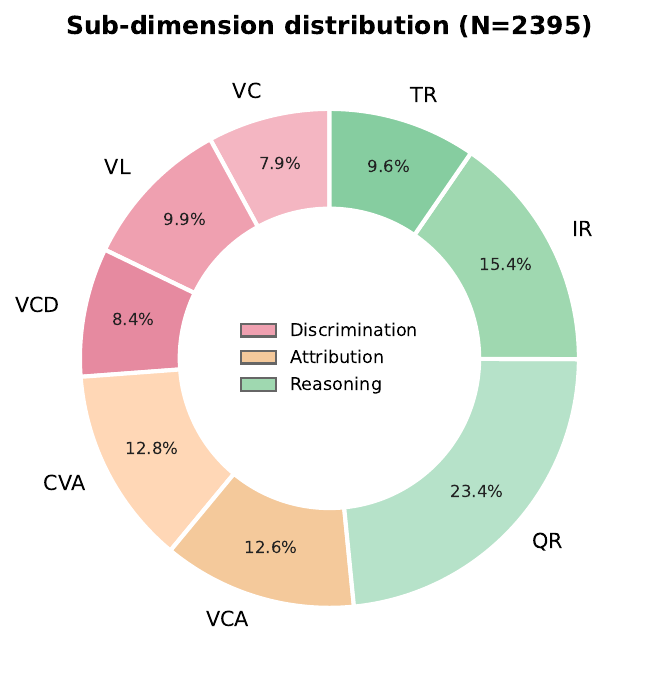}
\caption{Sub-dimension distribution of HEAR questions ($N\!=\!2{,}395$).  VC/VL/VCD make up Discrimination (top three slices), CVA/VCA make up Attribution, and QR/IR/TR make up Reasoning.}
\label{fig:subdim_donut}
\end{figure*}

\paragraph{Sub-dimension breakdown.}
Figure~\ref{fig:subdim_donut} unrolls the three top-level dimensions into the eight sub-dimensions that the benchmark scores.  Discrimination splits into VC ($7.9\%$), VL ($9.9\%$), and VCD ($8.4\%$); Attribution splits into CVA ($12.8\%$) and VCA ($12.6\%$); Reasoning splits into QR ($23.4\%$, the dominant Reasoning sub-type), IR ($15.4\%$), and TR ($9.6\%$).

\paragraph{VoxMM category coverage.}
Figure~\ref{fig:voxmm_categories} reports the genre composition of the VoxMM slice: \emph{Remote} ($104$ clips) and \emph{Documentary} ($72$) are the two heads, followed by \emph{Entertainment}, \emph{Commercial},
\emph{Interview}, and \emph{News} ($32$--$40$ each), with \emph{Politics}, \emph{Conversation}, \emph{Lecture}, \emph{Presentation}, and \emph{Sports} forming a long tail.

\begin{figure*}[t]
\centering
\includegraphics[width=\linewidth]{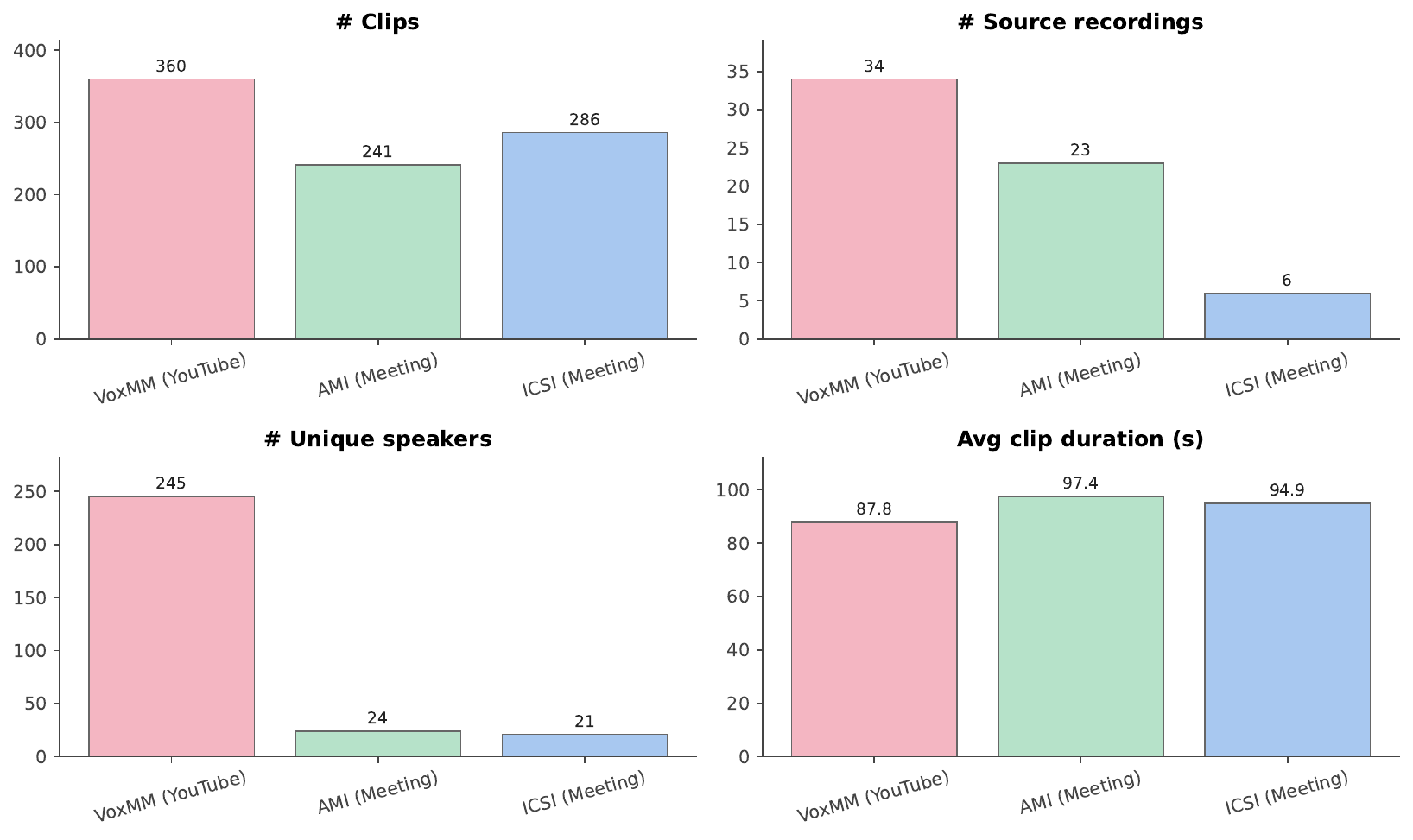}
\caption{Per-domain summary statistics for HEAR.}
\label{fig:per_domain}
\end{figure*}

\begin{figure*}[t]
\centering
\includegraphics[width=\linewidth]{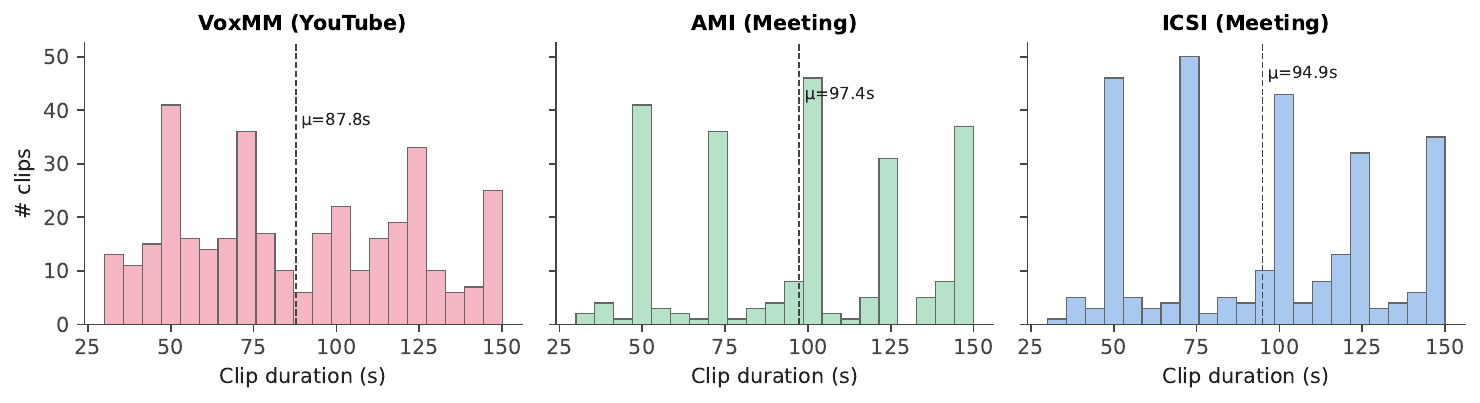}
\caption{Statistics of clip duration for HEAR.}
\label{fig:distributions}
\end{figure*}

\paragraph{Per-domain summary and clip-level distributions.}
Figure~\ref{fig:per_domain} consolidates four headline statistics across the three domains: clip count, distinct source recordings, distinct speakers, and average clip duration. Clip durations are evenly distributed as described in Figure~\ref{fig:distributions}.

\begin{figure*}[t]
\centering
\taskcell{accentDiscr}{VC}{Voice Counting}
  {\audiobadge{Main Audio}}
  {Listen to the main audio and answer the following multiple-choice question.\\[4pt]
   \textbf{How many distinct speakers are present in this audio clip?}}
  {Options: (A)\,2\quad (B)\,3\quad (C)\,4\quad (D)\,5\quad (E)\,6}

\vspace{10pt}

\taskcell{accentDiscr}{VL}{Voice Localisation}
  {\audiobadge{Main Audio + Reference Voice}}
  {Listen to the main audio, followed by a reference voice of the target speaker...\\[4pt]
   \textbf{Given the reference voice, select the time range where this speaker appears.}}
  {Options A--E are time ranges, e.g., [3.0s\,--\,5.5s].}

\vspace{10pt}

\taskcell{accentDiscr}{VL}{Voice Localisation}
  {\audiobadge{Main Audio + Reference Voice}}
  {Listen to the main audio, followed by a reference voice...\\[4pt]
   \textbf{Given the reference voice, select the time range where this speaker does NOT appear.}}
  {Options A--E are time ranges.}

\vspace{10pt}

\taskcell{accentDiscr}{VCD}{Voice-Change Detection}
  {\audiobadge{Main Audio}}
  {Listen to the main audio and answer the following multiple-choice question.\\[4pt]
   \textbf{In which time range does a speaker change (voice transition) occur?}}
  {Options A--E are time ranges.}

\vspace{10pt}

\taskcell{accentDiscr}{VCD}{Voice-Change Detection}
  {\audiobadge{Main Audio}}
  {Listen to the main audio and answer the following multiple-choice question.\\[4pt]
   \textbf{In which time range does speaker overlap (multiple speakers talking simultaneously) occur?}}
  {Options A--E are time ranges.}

\caption{Sample queries of \textit{Discrimination} dimension in HEAR benchmark.}
\label{fig:hear-v3-discrimination}
\end{figure*}

\begin{figure*}[t]
\centering
\taskcell{accentAttr}{CVA}{Content-to-Voice Attribution}
  {\audiobadge{Main Audio + Voice-Options (A--E)}}
  {Listen to the main audio, followed by the voice options audio...\\[4pt]
   \textbf{Which speaker said: \slot{X}?}}
  {Options (A)--(E) correspond to the voice options.}
\vspace{10pt}

\taskcell{accentAttr}{VCA}{Voice-to-Content Attribution}
  {\audiobadge{Main Audio + Reference Voice}}
  {Listen to the main audio, followed by a reference voice of the target speaker...\\[4pt]
   \textbf{Which of the following utterances was spoken by the given voice?}}
  {(A) \slotmath{X_1}\quad (B) \slotmath{X_2}\quad (C) \slotmath{X_3}\quad (D) \slotmath{X_4}\quad (E) \slotmath{X_5}}

\caption{Sample queries of \textit{Attribution} dimension in HEAR benchmark.}
\label{fig:hear-v3-attribution}
\end{figure*}

\begin{figure*}[t]
\centering
\taskcell{accentReas}{QR1}{Quantitative Reasoning}
{\audiobadge{Main Audio}}
{Listen to the main audio and answer the following multiple-choice question.\\[4pt]
\textbf{How many speakers mentioned \slot{X}?}}
{Options: (A)\,1\quad (B)\,2\quad (C)\,3\quad (D)\,4\quad (E)\,5}

\vspace{6pt}

\taskcell{accentReas}{QR2}{Quantitative Reasoning}
{\audiobadge{Main Audio}}
{Listen to the main audio and answer the following multiple-choice question.\\[4pt]
\textbf{Which of the following was said by the speaker with the $k$-th longest total speaking time?}}
{(A) \slotmath{X_1}\quad (B) \slotmath{X_2}\quad (C) \slotmath{X_3}\quad (D) \slotmath{X_4}\quad (E) \slotmath{X_5}}

\vspace{6pt}

\caption{Sample queries of \textit{Reasoning} dimension in HEAR benchmark.}
\label{fig:hear-v3-reasoning-a}
\end{figure*}

\begin{figure*}[t]
\centering
\taskcell{accentReas}{TR1}{Temporal Reasoning}
  {\audiobadge{Main Audio}}
  {Listen to the main audio and answer the following multiple-choice question.\\[4pt]
   \textbf{What was the very first utterance spoken by the person who said \slot{Y}?}}
  {Options A--E are candidate utterances.}

\vspace{6pt}

\taskcell{accentReas}{TR1}{Temporal Reasoning}
  {\audiobadge{Main Audio}}
  {Listen to the main audio and answer the following multiple-choice question.\\[4pt]
   \textbf{What was the very last utterance spoken by the person who said \slot{Y}?}}
  {Options A--E are candidate utterances.}

\taskcell{accentReas}{TR2}{Temporal Reasoning}
  {\audiobadge{Main Audio}}
  {Listen to the main audio and answer the following multiple-choice question.\\[4pt]
   \textbf{After someone said \slot{X}, what was the first thing said by the person who said \slot{Y}?}}
  {Options A--E are candidate utterances.}

\vspace{6pt}

\taskcell{accentReas}{IR}{Identity Reasoning}
  {\audiobadge{Main Audio}}
  {Listen to the main audio and answer the following multiple-choice question.\\[4pt]
   \textbf{Which of the following was also said by the person who said \slot{Y}?}}
  {Options A--E are candidate utterances.}

\caption{Sample queries of \textit{Reasoning} dimension in HEAR benchmark.}
\label{fig:hear-v3-reasoning-b}
\end{figure*}

\section{Details of Benchmarks Where Speaker Attribution Is Essential}
\label{app:benchmark_details}

We provide additional details on the construction and statistics of the three voice-dependent benchmarks used in our evaluation: WDYL (\emph{What Do You Like?}), GAOKAO, and FTS (\emph{Find the Spy}). 
All three benchmarks are designed such that correctly answering the question requires not only understanding the spoken content, but also identifying and tracking \emph{who} is speaking.

\paragraph{WDYL (What Do You Like?).}
WDYL evaluates whether a model can resolve first-person references using speaker identity.
Each example contains a conversation between two speakers, followed by a first-person question such as ``What is my favorite city?'' spoken in the cloned voice of one of the two speakers.
Because the referent of ``my'' depends entirely on the identity of the question speaker, the correct answer cannot be determined from the linguistic content of the question alone.

The benchmark contains 793 questions over 793 audio clips, totaling 2.36 hours of audio.
The average clip duration is 10.7\,s, with a range of 8.1--13.2\,s.
Each question has three answer choices: one corresponding to each speaker's stated preference and one distractor not mentioned by either speaker.
Since the distractor can be eliminated from the transcript alone, the effective chance accuracy is 50\%; accordingly, we use 50\% as the random-choice baseline in Table~\ref{tab:ood}.
The dialogue audio is taken from \cite{wu2024justasrllm}.
For each example, we append a 0.3\,s pause followed by a question synthesized with IndexTTS 2, using a cloned voice from one of the speakers in the preceding dialogue.

We initially synthesized 1,000 examples and retained 793 after speaker-similarity quality control using ECAPA embeddings, requiring a speaker embedding cosine similarity (SECS) of at least 0.70.
The retained examples have a mean SECS of 0.778.
The cloned question voice corresponds to the \emph{asker} in 476 examples and the \emph{answerer} in 317 examples.

\paragraph{GAOKAO.}
GAOKAO tests speaker-conditioned interpretation in natural English listening-comprehension conversations.
Each example consists of a short English dialogue followed by a first-person question synthesized in the cloned voice of one of the speakers.

The benchmark contains 93 questions over 93 clips, totaling 0.52 hours (30.9 minutes) of audio.
The average duration is 20.0\,s, ranging from 12.0\,s to 29.9\,s.
Each question has two answer choices, corresponding to a random-guess accuracy of 50\%.
The source audio consists of Gaokao English short-conversation recordings from the ICQ subset of \cite{hu2024wavllm}.
We append a 0.35\,s pause and an IndexTTS 2-generated question whose voice is cloned from a speech segment of the target speaker within the same dialogue.

We synthesized 97 examples and retained 93 after quality control.
Examples were required to achieve an ECAPA speaker-similarity score of at least 0.7 and a Whisper word error rate (WER) of at most 0.20.
The target speaker is male in 46 examples and female in 47 examples.
Gold-answer positions are distributed as A: 40 and B: 53.

\paragraph{FTS (Find the Spy).}
FTS evaluates multi-speaker voice tracking in a social-deduction setting.
Each example is a 3--6-player Find-the-Spy game in which every player receives the same secret word except for one designated spy.
The model observes the players' spoken clues and must identify which participant is the spy.

Crucially, players first introduce themselves by name, after which they provide their clues in a \emph{different, randomly shuffled order} and without explicitly stating their names.
Consequently, associating each clue with the corresponding player requires tracking speaker identity across temporally separated turns; textual content alone does not reveal which player produced which clue.

The benchmark contains 400 questions over 400 clips, totaling 3.67 hours of audio.
The mean duration is 33.1\,s, with clips ranging from 17.2\,s to 56.6\,s.
There are 100 games for each player count from three to six, yielding chance accuracies of 33.3\%, 25.0\%, 20.0\%, and 16.7\%, respectively.

Game configurations and clues are generated using the SocialMaze~\cite{xu2025socialmazebenchmarkevaluatingsocial} framework.
Each player's utterances are synthesized with IndexTTS 2, while keeping a single reference voice fixed for that player throughout the game.
\section{Experiment Details}
\label{appendix:baseline_hparams}
As introduced in Section~\ref{experiments}, we evaluate 20 leading SLMs categorized into three distinct families. The detailed list of models and their references is as follows:
\begin{itemize}
    \item \textbf{Omni-modality models}: MiniCPM-o series~\cite{yao2024minicpm, cui2026minicpmo45realtimefullduplex}, Qwen2.5-Omni series~\cite{xu2025qwen25omnitechnicalreport}, Qwen3-Omni-30B-A3B series~\cite{xu2025qwen3omnitechnicalreport}, Gemma-4-E4B-it~\cite{gemmateam2026gemma4technicalreport}, and Phi-4-Multimodal~\cite{microsoft2025phi4minitechnicalreportcompact}.
    \item \textbf{Spoken or Audio language models}: MiDashengLM~\cite{dinkel2026midashenglmefficientaudiounderstanding}, Voxtral~\cite{liu2025voxtral}, Fun-Audio-Chat~\cite{tongyifunteam2026funaudiochattechnicalreport}, Kimi-Audio~\cite{kimiteam2025kimiaudiotechnicalreport}, Qwen2-Audio~\cite{chu2024qwen2audiotechnicalreport}, Audio-Flamingo~3~\cite{goel2025audioflamingo3advancing}, and Step-Audio series~\cite{tian2025stepaudior1technicalreport, wu2025stepaudio2technicalreport}.
    \item \textbf{Proprietary models}: Gemini-3 series~\cite{gemini31pro2026, gemini3flash2026} and GPT-4o-audio-preview~\cite{openai2024gpt4o}.
\end{itemize}
For every open-source baseline, we adopt the inference hyperparameters specified in its official repository; for proprietary systems, we use the default API settings. The concrete experimental settings are comprehensively described in Table~\ref{tab:baseline_hparams}. All open-weight baselines are served through vLLM or the Hugging Face inference path, while the three closed-source endpoints (\textit{gpt-4o-audio-preview}, \textit{gemini-3-flash-preview}, \textit{gemini-3.1-pro-preview}) are queried with default setting. We report the value used in our configuration for reproducibility. Table~\ref{tab:baseline_hparams} summarizes the per-model defaults.
\section{Reward Composition Details}
\label{app:reward-details}

This appendix details the coefficients of the composite reward (Section~\ref{subsec:reward}) and the dynamic-programming implementation of the speaker-order reward $R_{\mathrm{ord}}$.

\subsection{Per-term Coefficients}

The five reward terms are linearly combined with fixed coefficients $\lambda_{x}$:
\begin{equation}
\small
\begin{aligned}
R(Y_i;\tau^\star,a^\star) &= \lambda_{\mathrm{ans}} R_{\mathrm{ans}}(\hat{a}_i;a^\star) \\[-2pt]
&\quad + \lambda_{\mathrm{fmt}} R_{\mathrm{fmt}}(Y_i) + \lambda_{\mathrm{cp}} R_{\mathrm{cp}}(\hat{\tau}_i;\tau^\star) \\[-2pt]
&\quad + \lambda_{\mathrm{cnt}} R_{\mathrm{cnt}}(\hat{\tau}_i;\tau^\star) + \lambda_{\mathrm{ord}} R_{\mathrm{ord}}(\hat{\tau}_i;\tau^\star).
\end{aligned}
\end{equation}

Table~\ref{tab:reward-coef} lists the coefficients used in our experiments.

\begin{table}[h]
\centering
\resizebox{\columnwidth}{!}{%
\begin{tabular}{lcc}
\toprule
Reward term & Symbol & Coefficient $\lambda$ \\
\midrule
Answer correctness & $R_{\mathrm{ans}}$ & $1.0$ \\
Response format & $R_{\mathrm{fmt}}$ & $0.2$ \\
Speaker-tagged transcript (cpWER)& $R_{\mathrm{cp}}$  & $0.5$ \\
Speaker-count & $R_{\mathrm{cnt}}$ & $0.5$ \\
Speaker-order (LCS) & $R_{\mathrm{ord}}$ & $0.5$ \\
\bottomrule
\end{tabular}%
}
\caption{Coefficients for the composite reward terms.}
\label{tab:reward-coef}
\end{table}

\subsection{Speaker-Order Reward via LCS}
\label{app:lcs}

To compute $R_{\mathrm{ord}}$, we extract the sequence of speaker labels in order of appearance from both the ground-truth ($Q=(q_1,\dots,q_m)$) and the prediction ($P=(p_1,\dots,p_n)$).

We define $R_{\mathrm{ord}}$ as the length-normalized Longest Common Subsequence (LCS):
\begin{equation}
R_{\mathrm{ord}}(\hat{\tau}_i;\tau^\star) = \frac{\mathrm{LCS}(P,Q)}{\max(|P|,|Q|,1)} \in [0,1].
\end{equation}

\begin{algorithm}[t]
\caption{Speaker-Order Reward $R_{\mathrm{ord}}$ (Length-normalized LCS)}
\label{alg:lcs-reward}
\small
\begin{algorithmic}[1]
\Require Predicted speaker labels $P=(p_1,\dots,p_n)$; GT speaker labels $Q=(q_1,\dots,q_m)$
\Ensure Reward $r\in[0,1]$
\State $D[i,0]\gets 0,\ D[0,j]\gets 0\quad \forall i,j$
\For{$i=1$ \textbf{to} $n$}
  \For{$j=1$ \textbf{to} $m$}
    \If{$p_i = q_j$}
      \State $D[i,j]\gets D[i\!-\!1,j\!-\!1]+1$
    \Else
      \State $D[i,j]\gets \max\!\bigl(D[i\!-\!1,j],\,D[i,j\!-\!1]\bigr)$
    \EndIf
  \EndFor
\EndFor
\State \Return $D[n,m]\,/\,\max(n,m,1)$
\end{algorithmic}
\end{algorithm}

$\mathrm{LCS}(P,Q)$ represents the longest sequence appearing in both $P$ and $Q$ in the same relative order.
Algorithm~\ref{alg:lcs-reward} outlines the $O(nm)$ dynamic-programming procedure.

\paragraph{From DP to Speaker Turns.}
The LCS algorithm directly aligns predicted ($P$) and ground-truth ($Q$) speaker sequences. When $p_i = q_j$, the DP matches a correctly attributed turn in temporal order ($+1$). When $p_i \neq q_j$, the $\max$ operation resolves misalignments by skipping either a hallucinated prediction or a missing GT turn. This ensures that isolated errors do not disrupt the global turn alignment.

\paragraph{Example.}
Suppose the maximum reward $R_{\mathrm{ord}}$ is normalized to 1 in this example. Given a 5-turn dialogue $Q=(s_1,s_2,s_1,s_3,s_2)$, our metric gracefully penalizes conversational errors. 
\emph{Case 1 (misattribution):} Predicting $P_1=(s_1,s_2,\mathbf{s_2},s_3,s_2)$ yields $\mathrm{LCS}=4$ and $R_{\mathrm{ord}}=4/5=0.8$. 
\emph{Case 2 (insertion):} Emitting an extra turn ($|P_2|=6$) keeps $\mathrm{LCS}=5$ but increases the denominator, yielding $R_{\mathrm{ord}}=5/\max(6,5)\approx 0.83$. 
\emph{Case 3 (deletion):} Missing a turn ($|P_3|=4$) yields $R_{\mathrm{ord}}=4/5=0.8$. 
Thus, misattributions, insertions, and deletions all incur proportional penalties, providing a robust turn-alignment signal.
\section{Training Details}
\label{app:training_details}

We fine-tune Qwen3-Omni-30B-A3B-Instruct with GRPO on $4{\times}$NVIDIA H200 GPUs. LoRA adapters are applied to all language-model linear projections with rank $r{=}64$, $\alpha{=}128$, and dropout $0$, while the audio/vision encoders and modality aligners remain frozen. Expert parallelism (EP$=$4) is used to fit the MoE backbone across GPUs.

Training runs for $1$ epoch with an effective batch size of $64$, formed by $16$ unique prompts and $4$ GRPO rollouts per prompt. We use bfloat16 precision and a learning rate of $5{\times}10^{-6}$ with cosine scheduling, $239$ warmup iterations, and cosine decay to $5{\times}10^{-7}$. Rollouts are generated with vLLM~\cite{kwon2023efficient} at temperature $1.0$ using sequence-level importance sampling, asymmetric PPO clipping ($\varepsilon_{\text{low}}{=}3{\times}10^{-4}$, $\varepsilon_{\text{high}}{=}4{\times}10^{-4}$), and $\beta{=}0$ KL regularization. Training is implemented on top of the \textsc{Megatron-SWIFT}~\cite{zhao2024swiftascalablelightweightinfrastructure} backend, which provides the expert-parallel MoE training support required for the Qwen3-Omni-30B-A3B backbone.

\section{Descriptions of VoiceBench}
\label{app:voicebench}
To evaluate the dyadic conversational capabilities of our fine-tuned A2R models, we utilize VoiceBench~\cite{chen2024voicebenchbenchmarkingllmbasedvoice}, a comprehensive benchmark featuring both real and synthetic spoken instructions. Following standard protocols, AlpacaEval, CommonEval, and WildVoice samples are assessed using gpt-5.4-mini~\cite{openai_gpt54mini_2026}. We evaluate our models across five key dimensions:

\paragraph{Open-Ended QA (AlpacaEval~\cite{dubois2025lengthcontrolledalpacaevalsimpleway}, CommonEval~\cite{ardila2020commonvoicemassivelymultilingualspeech}, WildVoice)}
assesses general knowledge and conversational capability. While \textbf{AlpacaEval} uses synthetic instructions to measure response quality, \textbf{CommonEval} and \textbf{WildVoice} (1,000 human-recorded samples) evaluate information-seeking capabilities under diverse, real-world human speech conditions.

\paragraph{Reference-Based QA (BBH)~\cite{suzgun2022challengingbigbenchtaskschainofthought}}
measures reasoning accuracy against reference answers. We use a human-recorded spoken version of \textbf{BBH} (BIG-Bench Hard) to evaluate multi-step reasoning (e.g., logical, arithmetic, and temporal) on long and complex real spoken instructions.

\paragraph{Multiple-Choice QA (MMSU~\cite{wang2026mmsumassivemultitaskspoken}, OpenBookQA) ~\cite{mihaylov2018suitarmorconductelectricity}}
evaluates general knowledge, elementary science, and common-sense reasoning by requiring models to select the correct option from a predefined set.

\paragraph{Instruction Following (IFEval)~\cite{gao2025ifevalaudiobenchmarkinginstructionfollowingcapability}}
tests the model's strict adherence to specific structural and formatting constraints using synthetic spoken instructions.

\paragraph{Safety (AdvBench)~\cite{zou2023universaltransferableadversarialattacks}}
measures the model's refusal rate against malicious prompts using synthetic speech, operating on the premise that a safe assistant must firmly reject harmful instructions.
\section{Paired Reasoning Accuracy}
\label{app:paired_reasoning}

For the reasoning axis, we evaluate each original clip jointly with its
semantic-hallucination counterpart. Let $c_i$ and $\tilde{c}_i$ denote
binary correctness indicators for the original and hallucinated clips in
pair $i$, respectively. We define paired reasoning accuracy as
\begin{equation}
R_{\mathrm{pair}}
= \frac{1}{N_R}\sum_{i=1}^{N_R} c_i \tilde{c}_i,
\end{equation}
where $N_R$ is the number of reasoning clip pairs. Thus, each pair
constitutes a single evaluation unit and contributes one correct
prediction only when both clips are answered correctly.

Discrimination and attribution are evaluated with standard accuracy,
whereas reasoning uses $R_{\mathrm{pair}}$. To account for the different
numbers of evaluation units across the three axes, we compute the overall
average by weighting each axis by its corresponding number of evaluation
units:
\begin{equation}
\mathrm{Avg.}
= \frac{
N_D D + N_A A + N_R R_{\mathrm{pair}}
}{
N_D + N_A + N_R
},
\end{equation}
where $N_D$ and $N_A$ denote the numbers of discrimination and attribution
examples, respectively, and $N_R$ denotes the number of reasoning clip
pairs. In particular, each reasoning pair is counted as one evaluation
unit rather than as two individual clips.

\section{Licenses and Intended Use}
We use all assets in accordance with their stated terms. AMI~\cite{10.1007/11677482_3}, ICSI~\cite{Janin2003TheIM}, and VoxMM~\cite{10446300} are released under CC BY 4.0. Our base model, Qwen3-Omni-30B-A3B-Instruct~\cite{xu2025qwen3omnitechnicalreport}, is released under Apache 2.0. 

\section{AI Tool Use}
We used AI-based writing assistance tools only for language editing, grammar correction, and improving the clarity of the manuscript.

\begin{table*}[p]
\centering\resizebox{\textwidth}{!}{
\setlength{\tabcolsep}{4pt}
\begin{tabular}{llccc}
\toprule
\textbf{Model} & \textbf{Checkpoint (HF / source)} & \textbf{Temp.} & \textbf{Top-$p$} & \textbf{Max tokens} \\
\midrule
Audio-Flamingo 3            & \texttt{nvidia/audio-flamingo-3-hf}          & 0.7 & 0.9   & 512   \\
Fun-Audio-Chat-8B           & \texttt{FunAudioLLM/Fun-Audio-Chat-8B}       & 0.6 & 0.95  & 512    \\
Gemma 4 E4B-it              & \texttt{google/gemma-4-E4B-it}               & 1.0 & 0.95  & 512   \\
Kimi-Audio-7B-Instruct & \texttt{moonshotai/Kimi-Audio-7B-Instruct} & 0.0 & --    & 2,048  \\
MiDashengLM-7B              & \texttt{mispeech/midashenglm-7b}             & 0.0 & --    & 512    \\
MiniCPM-o-2.6               & \texttt{openbmb/MiniCPM-o-2\_6}              & 0.3 & --    & 512   \\
MiniCPM-o-4.5               & \texttt{openbmb/MiniCPM-o-4\_5}              & 0.3 & 0.7   & 512   \\
Phi-4-Multimodal            & \texttt{microsoft/Phi-4-multimodal-instruct} & 0.0 & --    & 512    \\
Qwen2-Audio-7B-Instruct     & \texttt{Qwen/Qwen2-Audio-7B-Instruct}        & 0.7 & 0.5   & 512    \\
Qwen2.5-Omni-3B             & \texttt{Qwen/Qwen2.5-Omni-3B}                & 0.0 & --    & 512    \\
Qwen2.5-Omni-7B             & \texttt{Qwen/Qwen2.5-Omni-7B}                & 0.0 & --    & 512    \\
Qwen3-Omni-30B-A3B-Instruct & \texttt{Qwen/Qwen3-Omni-30B-A3B-Instruct}    & 0.6 & 0.95  & 8192  \\
Qwen3-Omni-30B-A3B-Thinking & \texttt{Qwen/Qwen3-Omni-30B-A3B-Thinking}    & 0.6 & 0.95  & 16,384 \\
Step-Audio-2-mini           & Step-Audio-2-mini (StepFun)                  & 0.1 & --    & 2,048    \\
Step-Audio-R1               & Step-Audio-R1 (StepFun, thinking mode)       & 0.7 & 0.9   & 16,384 \\
Voxtral-Mini-3B             & \texttt{mistralai/Voxtral-Mini-3B-2507}      & 0.2 & 0.95  & 512    \\
Voxtral-Small-24B           & \texttt{mistralai/Voxtral-Small-24B-2507}    & 0.2 & 0.95  & 1,024    \\
\bottomrule
\end{tabular}
}
\caption{\textbf{Decoding hyperparameters used for the open-weight baselines in our HEAR evaluation.} ``--'' denotes parameters left unset, using the vLLM engine’s default value (1.0)~\citep{kwon2023efficient}. For A2R, we use the same parameters as the Qwen3-Omni-30B-A3B-Instruct model.}
\label{tab:baseline_hparams}
\end{table*}


\definecolor{oursframe}{HTML}{1E3A8A}      
\definecolor{oursback}{HTML}{F8FAFC}       
\definecolor{ablationframe}{HTML}{4B5563}  
\definecolor{ablationback}{HTML}{F9FAFB}   

\tcbset{
    promptboxstyle/.style={
        enhanced,
        breakable,
        boxrule=1pt,
        arc=4pt,
        top=10pt,
        bottom=10pt,
        left=12pt,
        right=12pt,
        fonttitle=\bfseries,
        coltitle=white,
        attach boxed title to top left={yshift=-2mm, xshift=4mm},
        boxed title style={
            sharp corners=downhill, 
            arc=2pt, 
            boxrule=0pt,
            top=3pt, 
            bottom=3pt, 
            left=8pt, 
            right=8pt
        },
        shadow={2mm}{-1mm}{0mm}{gray!15!white}
    }
}


\begin{figure*}[!htbp]
    \centering

    \begin{tcolorbox}[
        colback=gray!5!white,
        colframe=black!70,
        boxrule=0.8pt,
        arc=3pt,
        left=8pt, right=8pt, top=10pt, bottom=10pt
    ]

    \begin{center}
        \textbf{\large A2R (Ours)}
    \end{center}

    \vspace{0.8em}

    Reason in two stages inside \texttt{<reasoning>...</reasoning>}. First, transcribe the multi-speaker conversation in the main audio inside \texttt{<transcript>...</transcript>} placed at the very start of \texttt{<reasoning>}. Use the format \texttt{`Speaker 1: ...\textbackslash nSpeaker 2: ...`}, labeling speakers in the order they first appear and joining each speaker's utterances on a single line in chronological order. Then continue reasoning grounded in that transcript, and finally provide the answer in JSON inside \texttt{<answer>...</answer>}.

    \vspace{0.8em}
    \noindent Example:

    \vspace{0.5em}
    \noindent\texttt{<reasoning><transcript>Speaker 1: ...\textbackslash nSpeaker 2: ...</transcript>} \\
    \texttt{...your detailed reasoning grounded in the transcript...</reasoning>}

    \vspace{0.5em}
    \noindent\texttt{<answer>\{"Answer": "(A)"\}</answer>}

    \end{tcolorbox}

    \vspace{1.2em}

    \begin{tcolorbox}[
        colback=gray!5!white,
        colframe=black!70,
        boxrule=0.8pt,
        arc=3pt,
        left=8pt, right=8pt, top=10pt, bottom=10pt
    ]

    \begin{center}
        \textbf{\large Ablation}
    \end{center}

    \vspace{0.8em}

    Reason briefly inside \texttt{<reasoning>...</reasoning>}, then output your answer in JSON inside \texttt{<answer>...</answer>}.

    \vspace{0.8em}
    \noindent Example:

    \vspace{0.5em}
    \noindent\texttt{<reasoning>Your brief reasoning here.</reasoning>}

    \vspace{0.5em}
    \noindent\texttt{<answer>\{"Answer": "(A)"\}</answer>}

    \end{tcolorbox}

    \caption{Instruction prompts used for GRPO experiments.}
    \label{fig:prompts}
\end{figure*}

\begin{figure*}[!htbp]
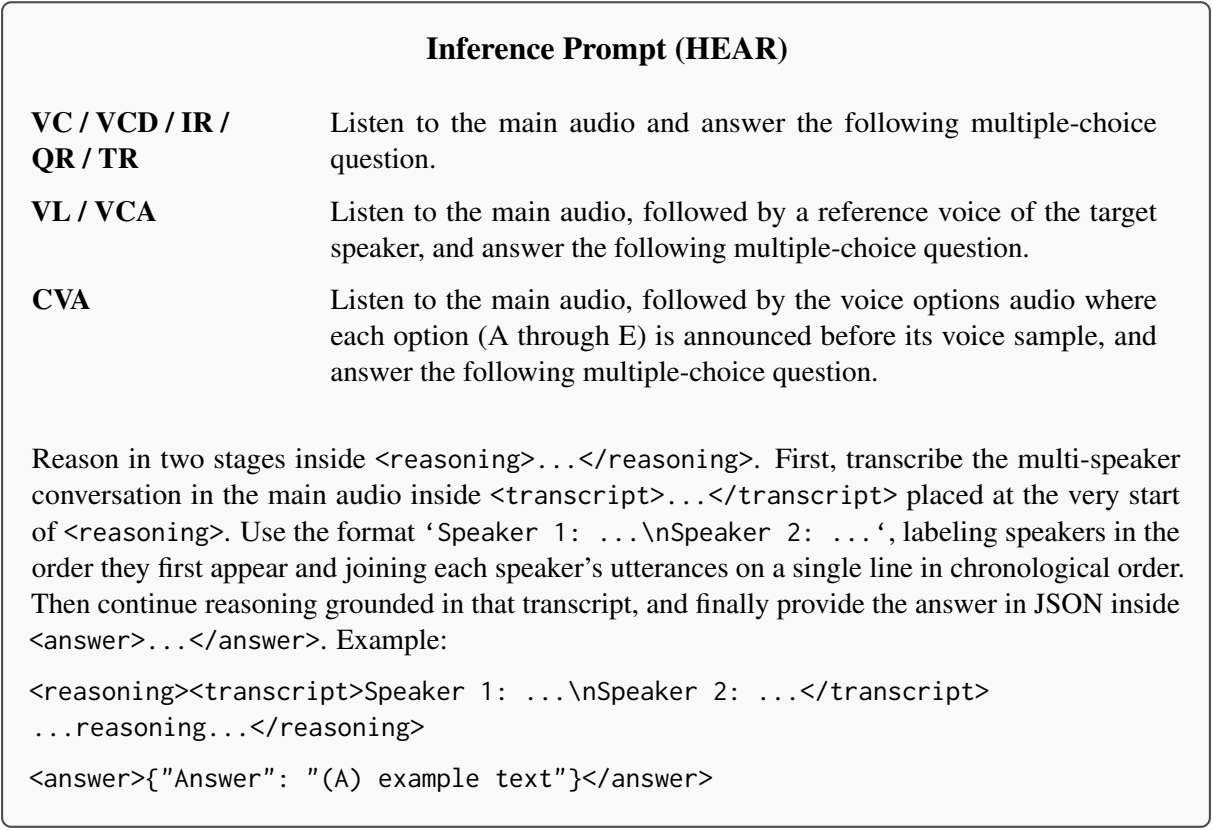

    \centering
    \begin{tcolorbox}[
        colback=gray!5!white,
        colframe=black!70,   
        boxrule=0.8pt,       
        arc=3pt,             
        left=8pt, right=8pt, top=10pt, bottom=10pt 
    ]
    
    \begin{center}
        \textbf{\large Inference Prompt (HEAR)}
    \end{center}
    
    \vspace{0.8em}
    
    \noindent
    \begin{minipage}[t]{0.26\linewidth}
        \textbf{VC / VCD / IR /} \\
        \textbf{QR / TR}
    \end{minipage}%
    \begin{minipage}[t]{0.72\linewidth}
        Listen to the main audio and answer the following multiple-choice question.
    \end{minipage}
    
    \vspace{0.8em}
    
    \noindent
    \begin{minipage}[t]{0.26\linewidth}
        \textbf{VL / VCA}
    \end{minipage}%
    \begin{minipage}[t]{0.72\linewidth}
        Listen to the main audio, followed by a reference voice of the target speaker, and answer the following multiple-choice question.
    \end{minipage}
    
    \vspace{0.8em}
    
    \noindent
    \begin{minipage}[t]{0.26\linewidth}
        \textbf{CVA}
    \end{minipage}%
    \begin{minipage}[t]{0.72\linewidth}
        Listen to the main audio, followed by the voice options audio where each option (A through E) is announced before its voice sample, and answer the following multiple-choice question.
    \end{minipage}

    \vspace{2em}
    
    Reason in two stages inside \texttt{<reasoning>...</reasoning>}. First, transcribe the multi-speaker conversation in the main audio inside \texttt{<transcript>...</transcript>} placed at the very start of \texttt{<reasoning>}. Use the format \texttt{`Speaker 1: ...\textbackslash nSpeaker 2: ...`}, labeling speakers in the order they first appear and joining each speaker's utterances on a single line in chronological order. Then continue reasoning grounded in that transcript, and finally provide the answer in JSON inside \texttt{<answer>...</answer>}. Example:
    
    \vspace{0.5em}
    \noindent\texttt{<reasoning><transcript>Speaker 1: ...\textbackslash nSpeaker 2: ...</transcript>} \\
    \texttt{...reasoning...</reasoning>}
    
    \vspace{0.5em}
    \noindent\texttt{<answer>\{"Answer": "(A) example text"\}</answer>}
    
    \end{tcolorbox}
    
    \caption{Instruction prompt template used for HEAR benchmark, consisting of a task-specific listening prefix and a common reasoning suffix.}
    \label{fig:prompts}
\end{figure*}


\begin{figure*}[!htbp]
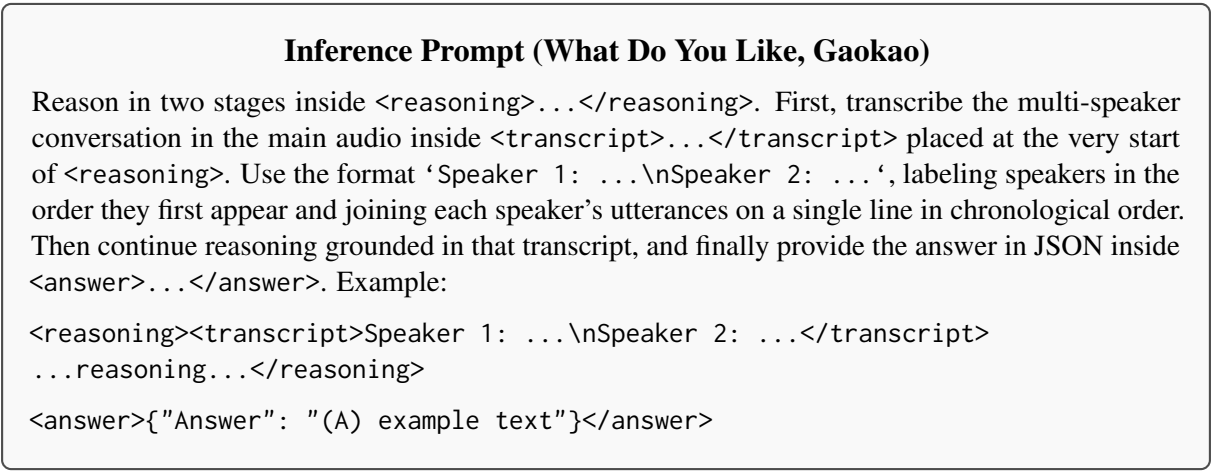

    \centering
    \begin{tcolorbox}[
        colback=gray!5!white, 
        colframe=black!70,    
        boxrule=0.8pt,        
        arc=3pt,              
        left=8pt, right=8pt, top=10pt, bottom=10pt 
    ]
    
    \begin{center}
        \textbf{\large Inference Prompt (What Do You Like, Gaokao)}
    \end{center}
    
    Reason in two stages inside \texttt{<reasoning>...</reasoning>}. First, transcribe the multi-speaker conversation in the main audio inside \texttt{<transcript>...</transcript>} placed at the very start of \texttt{<reasoning>}. Use the format \texttt{`Speaker 1: ...\textbackslash nSpeaker 2: ...`}, labeling speakers in the order they first appear and joining each speaker's utterances on a single line in chronological order. Then continue reasoning grounded in that transcript, and finally provide the answer in JSON inside \texttt{<answer>...</answer>}. Example:
    
    \vspace{0.5em}
    \noindent\texttt{<reasoning><transcript>Speaker 1: ...\textbackslash nSpeaker 2: ...</transcript>} \\
    \texttt{...reasoning...</reasoning>}
    
    \vspace{0.5em}
    \noindent\texttt{<answer>\{"Answer": "(A) example text"\}</answer>}
    
    \end{tcolorbox}
    
    \caption{Instruction prompt template used for What Do You Like and Gaokao benchmarks, consisting of a task-specific listening prefix and a common reasoning suffix.}
    \label{fig:prompts2}
\end{figure*}


\begin{figure*}[!htbp]
    \centering
    \begin{tcolorbox}[
        colback=gray!5!white, 
        colframe=black!70,  
        boxrule=0.8pt,       
        arc=3pt,              
        left=8pt, right=8pt, top=10pt, bottom=10pt 
    ]
    
    \begin{center}
        \textbf{\large Instruction Prompt (Find the SPY)}
    \end{center}
    
    \vspace{1em}
    
    \noindent
    You are playing the game ``Find the Spy.'' There are 3 players: \{Player Name 1\}, \{Player Name 2\} .... All players except one received the same word; one player, the spy, received a different word. Each player gives a short description of their word without saying it directly. Your task is to identify which player is describing a different word.
    
    \vspace{0.8em}
    \noindent Based on a recording of the game, decide which player is the spy.
    
    \vspace{0.8em}
    \noindent Options:\\
    (A) Rachel\\
    (B) Benjamin\\
    (C) Grace \\
    (D) ...

    \vspace{2em}
    
    Reason in two stages inside \texttt{<reasoning>...</reasoning>}. First, transcribe the multi-speaker conversation in the main audio inside \texttt{<transcript>...</transcript>} placed at the very start of \texttt{<reasoning>}. Use the format \texttt{`Speaker 1: ...\textbackslash nSpeaker 2: ...`}, labeling speakers in the order they first appear and joining each speaker's utterances on a single line in chronological order. Then continue reasoning grounded in that transcript, and finally provide the answer in JSON inside \texttt{<answer>...</answer>}. Example:
    
    \vspace{0.5em}
    \noindent\texttt{<reasoning><transcript>Speaker 1: ...\textbackslash nSpeaker 2: ...</transcript>} \\
    \texttt{...reasoning...</reasoning>}
    
    \vspace{0.5em}
    \noindent\texttt{<answer>\{"Answer": "(A) example text"\}</answer>}
    
    \end{tcolorbox}
    
    \caption{Instruction prompt used for the Find the Spy benchmark.}
    \label{fig:prompt_spy}
\end{figure*}


\end{document}